# PARAMANU: Compact and Competitive Monolingual Language Models for Low-Resource Morphologically Rich Indian Languages

**Mitodru Niyogi**[1] **Eric Gaussier**[1] **Arnab Bhattacharya**[2]
[1]Université Grenoble Alpes, CNRS, Grenoble INP, LIG, 38000 Grenoble, France
[2]Dept. of Computer Science and Engineering, Indian Institute of Technology Kanpur, Kanpur, India
mitodru.niyogi@cnrs.fr, eric.gaussier@imag.fr, arnabb@cse.iitk.ac.in

## Abstract

Multilingual large language models (LLMs) are expensive to pretrain and often suffer from imbalances across languages and datasets, English-centric bias, tokenizer oversegmentation for morphologically rich low-resource languages, and the curse of multilinguality. We introduce PARAMANU, the first family of Indian-only autoregressive language models trained from scratch on open-source language-specific data for the five most spoken Indian languages: Bengali, Hindi, Marathi, Tamil, and Telugu. All models are designed for *affordability* and are trained on a *single GPU* with a budget *under $1,000*, allowing under-resourced researchers to build competitive language models. To address low-resource challenges, we develop morphology-aligned, low-fertility *tokenizers*, propose an interpolation-based method for token position indices in RoPE based scaling to train longer sequences efficiently. We also create instruction-tuning datasets in Bangla that are translated to the other four languages. Despite their small size (108M-367M parameters), *Paramanu* achieves a strong performance-efficiency tradeoff and outperforms most larger multilingual models across all five languages. Our collection is available at https://huggingface.co/collections/mitodru/paramanu.

## 1 Introduction

Despite the existence of over 7,000 languages globally, current NLP and GenAI technologies remain heavily skewed towards English and other high-resource European languages, leaving a significant portion of the world's population, particularly speakers of global south languages, underserved (Schwartz, 2022; Nekoto et al., 2020; Choudhury, 2023). Indian languages, spoken by approximately 1.4 billion people, are among the most neglected, despite the fact that Hindi and Bangla are respectively the 5th and 6th most spoken[1] languages globally. Challenges such as lack of high-quality datasets, poor tokenization, and limited representation in pretraining corpora render Indian languages "low-resource" (Tsvetkov, 2017; Singh, 2008); being morphologically rich further impedes their performance (Joshi et al., 2020; Goyal et al., 2022; Nigatu et al., 2024).

Large language models (LLMs) like GPT-2 (Radford et al., 2019), LLaMa (Touvron et al., 2023), GPT-NeoX (Black et al., 2022), OPT (Zhang et al., 2022), Falcon (Almazrouei et al., 2023), and PaLM (Chowdhery et al., 2023) are predominantly trained on English and Latin-script languages, showing significantly degraded performance on Indian and other low-resource languages (Bang et al., 2023; Lai et al., 2023a). This disparity persists even in multilingual decoder-only LLMs (e.g., Bloom (Workshop et al., 2023), xGLM (Lin et al., 2022), mGPT (Shliazhko et al., 2024), Aya23 (Aryabumi et al., 2024), Llama-3 (Grattafiori et al., 2024), Llama-3.2 (Meta AI, 2024)). As these models have English-centric bias, and "think in English" (Schut et al., 2025; Guo et al., 2024a) often causes them to perform worse in non-Latin script languages (Shafayat et al., 2024; Shi et al., 2023; Huang et al., 2023; Bang et al., 2023). They also suffer from data and language-dependent imbalance (Dangarikar et al., 2024), a language-fits-all tokenizer resulting in bias, over-segmentation (Ahuja et al., 2023), unfair representation (Pfeiffer et al., 2021), language confusion (Marchisio et al., 2024) and reduced fluency (Guo et al., 2024b). This results in high token fertility for Indian languages, and increased inference and training costs.

Adapting existing LLMs to Indian languages through continual pretraining (Zheng et al., 2024a; Ji et al., 2025; Alves et al., 2024) and fine-tuning (Lu et al., 2024; Zheng et al., 2024b) presents mul-tiple challenges, including requirement of large data and compute, vocabulary extension inefficiencies, embedding alignment, and risk of catastrophic forgetting (Zheng et al., 2024a; Ahuja et al., 2023). Moreover, this adaptation pipeline assumes the suitability of English-centric foundations, which may not generalize well to Indian typologies and scripts.

[1]https://www.babbel.com/en/magazine/the-10-most-spoken-languages-in-the-world

We introduce *Paramanu*[2], the first family of Indian-only, openly licensed, non-commercial (CC BY-NC-SA 4.0) sub-400M decoder language models trained from scratch on fully open-source, language-specific Indian data. Paramanu comprises monolingual generative models for the five most spoken Indian languages: Bangla (Bengali), Hindi, Marathi, Tamil, and Telugu ranging from 108M to 367M parameters. We use low-fertility, language-specific tokenizers to reduce training and inference cost and latency. This approach avoids cross-language data imbalance, allows effective preprocessing of smaller corpora (including low-resource settings), and provides a controlled foundation for analyzing LLM behavior and developing downstream adaptations. We show that language-specific models achieve strong performance even under severe constraints[3] for low-resource, morphologically rich languages. Across a comprehensive evaluation against monolingual and multilingual models up to 8B parameters, *Paramanu* achieves an efficient performance-cost tradeoff, outperforming most larger models and, to our knowledge, all LLMs under 3B parameters.

We summarize our contributions as follows:

1. We present *Paramanu*, the first from-scratch, Indian-only open-source sub-400M decoder LMs for five major Indian languages, empirically demonstrating that small monolingual models can outperform much larger multilingual models under tight compute and budget constraints, making them broadly usable by NLP researchers working on Indian languages.
2. We developed morphology-aligned, low-fertility Indian tokenizers by combining Unigram tokens into BPE tokenizer.
3. We developed an interpolation method for the position indices of tokens in RoPE based scaling for training longer sequences on a single GPU.
4. We cleaned the training corpus and developed novel instruction-tuning datasets in Bangla, which were then machine translated and used for Hindi, Marathi, Tamil, and Telugu to further align our models with human instructions.

[2]Available at: https://huggingface.co/collections/mitodru/paramanu.

[3]Typically, a single GPU and <$1,000 budget for training.

The remainder of the paper is organized as follows: Section 2 discusses prior work; Section 3 details data, tokenization, and model design; Section 4 presents experiments, training, evaluations, and ablations; Section 5 presents discussion while Section 6 concludes.

## 2 Related Work

Multilingual large language models (LLMs) such as Bloom (Workshop et al., 2023), xGLM (Lin et al., 2022), and Sarvam 2B (Sarvam2B, 2024) have made significant progress in scaling decoder-only models across multiple languages. However, many of these models remain heavily English-biased: for instance, Llama-3.2 (Dubey et al., 2024) includes only 8% non-English tokens, while Sarvam 2B uses 40-50% English data. Such imbalances, together with tokenizer over-segmentation, disproportionately affect low-resource, morphologically rich languages. Efforts to adapt English-centric models for Indian languages include Airavata (Gala et al., 2024), OpenHathi (SarvamAI, 2023), Nemotron-Hindi (Joshi et al., 2025), and TamilLlama, which extend vocabularies and leverage fine-tuning techniques like LoRA (Hu et al., 2022) and QLoRA (Dettmers et al., 2023). Dedicated Indian-language LLMs trained from scratch-such as Aya23 8B (Aryabumi et al., 2024), mGPT (Shliazhko et al., 2024), and Sarvam 2B (Sarvam2B, 2024)still rely heavily on English data and struggle to generate high-quality text in Indian languages.

Massively multilingual models (MMTs) (Devlin et al., 2019; Conneau et al., 2020; Xue et al., 2021) are pretrained on large corpora across many languages but often lack alignment between distant languages, resulting in poor transfer performance (Lauscher et al., 2020). Studies attribute this to lower-quality tokenization per language (Rust et al., 2021) and show that adding multilingual data helps low-resource languages only until model capacity is reached, while consistently degrading performance for high-resource languages (Chang et al., 2024). Indic NLP research also suffers from a lack of culturally and linguistically relevant datasets (Doddapaneni et al., 2023; Khan et al., 2024), as most supervised datasets are translations from English. Recent efforts have explored building monolingual autoregressive LMs from scratch, such as German LLäMmlein (Pfister et al., 2025), BanglaT5 (Bhattacharjee et al., 2023), and BanglaByT5 (Bhattacharyya and Bhattacharya, 2025). Unlike BanglaByT5, which compared only models below 1B parameters, our work evaluates *Paramanu* against models up to 8B parameters, demonstrating the effectiveness of small, language-specific LLMs under resource constraints. This family of models can be used and extended by any NLP group working on Indian languages.

## 3 Methodology

We pretrain decoder-only generative language models from scratch under 400M parameters on Indian languages corpora of less than 15GB (25-66B tokens). In this section, we discuss details regarding our datasets, preprocessing, novel tokenization, multilingual instruction dataset creation, and context scaling with tokens positional interpolation.

### 3.1 Dataset for Pretraining

The pretrained data of 54.6 GB UTF-8 bytes for 5 major Indian languages was split into 95% training and 5% validation to retain as much data as possible, since the goal is to take a step toward developing effective pretrained generative language models from scratch for 5 major Indian languages. The pretraining corpus primarily consists of web-scraped news, blogs, and Wikipedia from IndianCorp v2 (Doddapaneni et al., 2023) for Marathi, Tamil, and Telugu, which was used to train IndicBERT-v2, and Bangla literature from Vacaspati (Bhattacharyya et al., 2023), used in training Bangla Electra. IndianCorp v2 also includes Indian language data from Wikipedia and OSCAR (Ortiz Suárez et al., 2019). Our pretraining corpora have no source code, scientific journals/articles, scientific books, legal and constitutional documents, etc. Dataset details can be found in Table 1.

### 3.2 Data Cleaning

Following prior work (Doddapaneni et al., 2023; Abadji et al., 2022), we perform regex-based filtering of HTML/XML tags, emails, links, emojis, personal info, and remove non-literal and foreign-script characters. Language identification is done using `cld3`[4] and IndianLID-FTN (Madhani et al., 2023) to discard non-target languages. We filter toxic content using Team et al. (2022), normalize whitespace and Unicode, and deduplicate paragraphs using 128-bit MurmurHash[5]. For Indian scripts (Bengali, Devanagari, Tamil, and Telugu), sentence splitting uses language-specific punctuation (danda "।" for Bengali and Devanagari scripts).

### 3.3 Tokenization

To improve morphological subword representations for Indian languages, we employ a hybrid tokenization approach that fuses vocabularies from independently trained SentencePiece models using Byte Pair Encoding (BPE; Sennrich et al., 2016) and the Unigram Language Model (Unigram LM; Kudo, 2018). This design is motivated by Bostrom and Durrett (2020), who show that Unigram LM produces subword units that better capture morphological structure through global optimization and probabilistic pruning, yielding cleaner subword inventories than greedy merge-based methods.

Both tokenizers are converted to SentencePieces `ModelProto` format, which serializes the vocabulary, subword scores, normalization rules, and special tokens, and the BPE vocabulary is augmented with all Unigram LM tokens not already present. The added tokens mainly correspond to productive prefixes, suffixes, and frequent short stems that are often split across multiple BPE merges. During tokenization, SentencePiece performs a Viterbi-style search over the combined vocabulary. Original BPE merges remain deterministic, while the added Unigram tokens are assigned scores from the Unigram LM, allowing them to compete with BPE merges. For many words, the resulting segmentation is identical to BPE alone (rows 4-5 in Table 7 in Appendix A), but for substrings where Unigram tokens better match common morphemes, the hybrid tokenizer selects these as atomic units, producing more morphologically coherent subwords.

Table 7 in Appendix A illustrates the tokenization behavior of our hybrid tokenizer compared to standard BPE across several Indian languages. In Telugu, the word ఆధారపడతాము (ādhārapaḍatāmu) is segmented by BPE as ['ఆధార', 'ప', 'డ', 'తాము'] (['ādhāra', 'pa', 'ḍa', 'tāmu']), which breaks the root ఆధారపడ into separate tokens. In contrast, our hybrid tokenizer produces ['ఆధారపడ', 'తాము'] (['ādhārapaḍa', 'tāmu']), preserving the root as a single unit and the suffix separately, maintaining the morphological and semantic structure of the word. For Tamil, the word பயணித்தார்கள் (payaṇittārkaḷ) is segmented by BPE as ['பயண', 'இத்தார்கள்'] (['payaṇa', 'ittārkaḷ']), splitting the verb root பயணி (travel) and the past-tense participle plus plural suffix into unnatural fragments. The hybrid tokenizer segments it as ['பயணித்த', 'ஆர்கள்'] (['payaṇitta', 'ārkaḷ']), keeping the root plus tense marker பயணித்த together and the plural marker ஆர்கள் as a separate token, which better reflects the underlying morphological units. Across these examples, the hybrid tokenizer consistently preserves stems and frequent suffixes as atomic subwords, whereas BPE often produces over-fragmented tokens. By combining BPE with Unigram tokens, it increases lexical coverage by representing both frequent and rare morphemes as reusable units. This enables more compact token sequences, reduces embedding redundancy, and generates subword representations that better align with the semantic and morphological structure of morphologically rich Indian languages.

[4] https://github.com/google/cld3
[5] https://pypi.org/project/mmh3/

| Language | Family | Script | Corpus Source | Corpus Size (GB) | #Sentences | #Speakers |
|---|---|---|---|---|---|---|
| Bangla | Indo-European | Bangla | Vacasapati + Wikipedia | 3.6 | 22,533,608 | 300 M |
| Hindi | Indo-European | Devanagari | IITB monolingual | 15.8 | 52,124,643 | 692 M |
| Marathi | Indo-European | Devanagari | Indian Corp v2 | 12.5 | 34,567,839 | 99 M |
| Tamil | Indo-Dravidian | Tamil | Indian Corp v2 | 10.7 | 27,872,768 | 77 M |
| Telugu | Indo-Dravidian | Telugu | Indian Corp v2 | 13.5 | 40,241,847 | 95 M |

Table 1: Pretraining data details after data cleaning along with language families, scripts, and speaker estimates. Speaker data is from the Indian Census 2011.

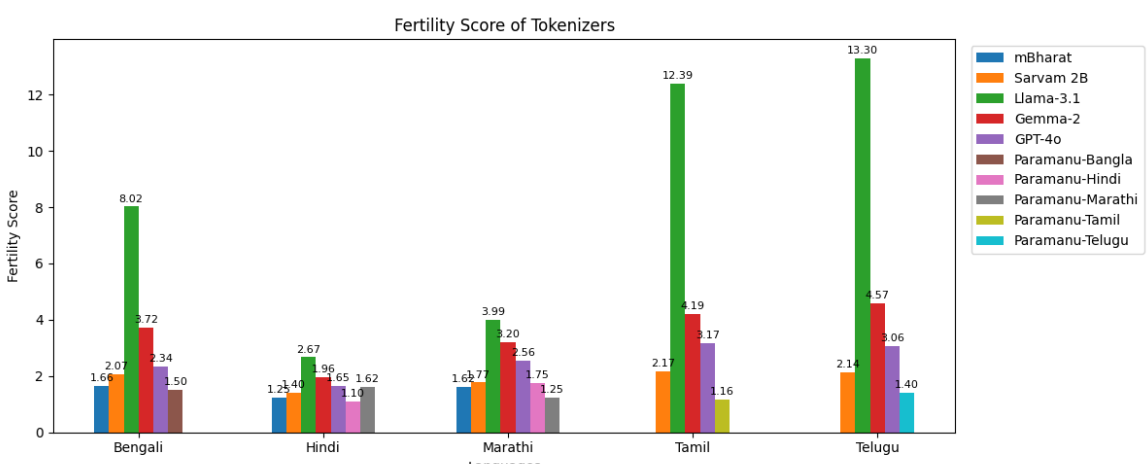

Figure 1: Fertility score of our tokenizer v/s LLMs across languages of 4 scripts (Bengali, Devanagari, Tamil, and Telugu). LLMs score are reported from (Sarvam2B, 2024).

| n_params | d_model | n_layers | n_heads | n_kv_heads | Seq Len |
|---|---|---|---|---|---|
| 108M | 768 | 12 | 12 | 12 | 1024 |
| 139M | 896 | 14 | 14 | 14 | 1024 |
| 162M | 1024 | 12 | 16 | 16 | 1024 |
| 208M | 1024 | 16 | 16 | 16 | 1024 |
| 237M | 1024 | 18 | 18 | 18 | 1024 |
| 367M | 1280 | 18 | 10 | 10 | 1024 |

Table 2: Model size configuration

During pre-tokenization, we apply NFC normalization, digit splitting, and byte fallback for unknown UTF-8 characters. Our tokenizers achieve the least fertility scores across all five languages compared to Sarvam 2B (Sarvam2B, 2024), Llama-3.1 (Dubey et al., 2024), Gemma-2 (Team et al., 2024), and GPT-4o (shown in Fig. 1).

### 3.4 Instruction Tuning Datasets

We constructed 23K instructions for Bangla from three sources: 5K human-authored instructions (on culture, literary, practical domain) by 20 native Bangla-speaking annotators, following guidelines more detailed in Appendix B, 15K translated instructions from Dolly (Conover et al., 2023), and 3K self-instruct-generated samples (Wang et al., 2023). These were translated to Hindi, Marathi, Tamil, and Telugu using Google Translate[6] with manual post-editing[7] The details of the dataset is described in Table 8 in the Appendix.

### 3.5 Context Scaling with RoPE Embeddings for Efficient Pretraining

We employ a scaled variant of Rotary Positional Embeddings (RoPE; Su et al., 2022) with a base value of $\theta = 10{,}000$. Inspired from Chen et al. (2023) to support pretraining with large context lengths on hardware-constrained settings (e.g., a single A100 40GB GPU), we introduce a *shrinking factor* that scales the input token position ids before the RoPE methodology is applied. This *shrinking factor* is defined as the ratio of the target context length $y$ to a fixed *permissible_context_size_length* $L$, which corresponds

[6] https://cloud.google.com/

[7] Available at: https://huggingface.co/collections/mitodru/paramanu.

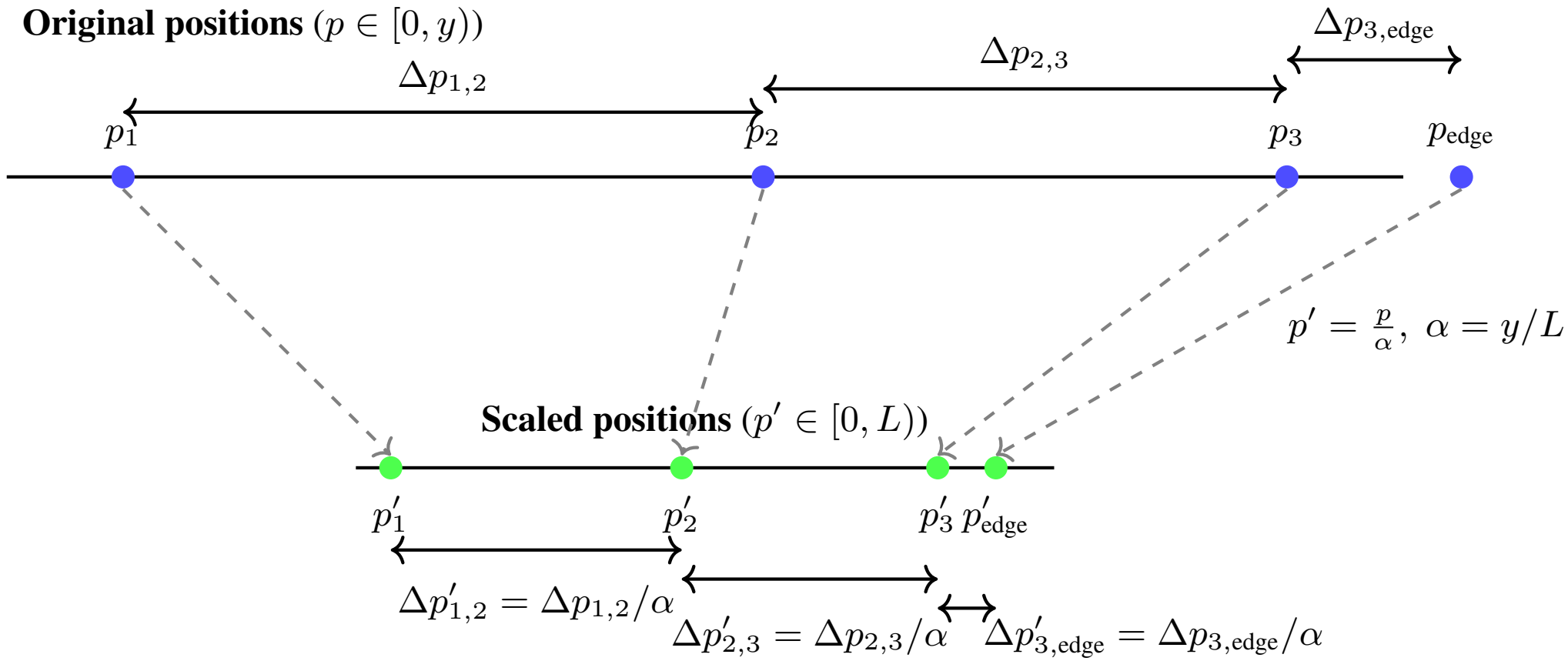

**RoPE applied after scaling**

Fractional positions are valid; attention depends only on relative distances to the maximum context length that the available hardware can accommodate. All other training hyperparameters such as batch size, vocabulary size, and model dimensions remain unchanged. Formally, for each token position $p$, we compute a scaled position $p'$ as:

$$p' = \frac{p}{\alpha} = \frac{p \cdot L}{y}$$

Figure 2: RoPE context scaling via positional interpolation. Tokens in the original context (blue) are linearly mapped to a hardware-supported range (green). Absolute distances shrink, relative offsets $\Delta p' = \Delta p/\alpha$ are preserved. Brackets for $\Delta p'$ are spaced to avoid overlap with token labels.

For example, with a target context length of $y = 4096$ and a permissible length of $L = 256$, the shrinking factor is $\alpha = \frac{4096}{256} = 16$. A token at position $p = 4000$ is mapped to $p' = \frac{4000}{16} = 250.00$, and its neighbor at $p = 4001$ maps to $p' \approx 250.06$. This ensures all scaled positions lie within the permissible range $[0, L]$ This is how we can capture higher context size during pretraining outside the *permissible_context_size_length*. Further, exploiting the fact that positional embeddings can be applied to non-integer positions as using RoPE, the self-attention score is only dependent on the relative position of the tokens through trigonometric functions. This scaling preserves the models ability to learn long-range dependencies. Fig. 2 illustrates the methodology.

## 4 Experiments

Our monolingual PARAMANU models are based on transformer (Vaswani et al., 2017) causal decoder architecture (Radford et al., 2019). Following Chinchilla (Hoffmann et al., 2022a), LLaMa (Touvron et al., 2023), we used RMSNorm (Zhang and Sennrich, 2019) with norm_epsilon =1e-5, SwiGLU (Shazeer, 2020) activation function, and an activation hidden size of $\sim \frac{8}{3}d$. Following (Chowdhery et al., 2023), we removed all biases from dense layers to improve the training stability. We also used weight tying (Press and Wolf, 2017) to improve the performance of language models by sharing the weights of the embedding and softmax layers. We followed Chinchilla scaling laws and the GPT-2 ratio (a 1.5B model pretrained on 40 GB of text data) (Radford et al., 2019) to estimate compute optimal model parameter configuration as shown in the Table 2 for our models given the size of the dataset.

### 4.1 Training

We pretrained our models using the AdamW optimizer (Loshchilov and Hutter, 2019), with $\beta_1 = 0.9$, $\beta_2 = 0.95$, eps $= 10^{-5}$. We use a cosine learning rate schedule, with warmup of 1000 steps, and decay final learning rate down to 10% of the peak learning rate. We use a weight decay of 0.1 and gradient clipping of 1.0. To further speedup training, we also used BF16 mixed precision training. We performed hyperparameter tuning on 15M models and used the concept of $\mu P$ transfer (Yang et al., 2021) to transfer the learned hyperparameters to our bigger models. All models are pretrained for 100K training steps except Hindi 367M (150K). For instruction-tuning, we followed Taori et al. (2023). Further details are in Appendix D.

| Model | Size | #langs | Bangla | Devanagari | | Tamil | Telugu | Training | Context | Pretrained Tokens | |
|---|---|---|---|---|---|---|---|---|---|---|---|
| | | | | Marathi | Hindi | | | Hours | Size | Total | Indian |
| Paramanu-Bangla | 108M | 1 | 25.22 | - | - | - | - | 42.75 | 1024 | 26.21B | 26.21B |
| Paramanu-Bangla-instruct | 108M | 1 | **29.52** | - | - | - | - | +0.5 | 1024 | +3.5M | +3.5M |
| Paramanu-Marathi | 208M | 1 | - | 26.40 | 30.97 | - | - | 88 | 1024 | 28.83B | 28.83B |
| Paramanu-Marathi-instruct | 208M | 1 | - | **26.93** | 30.54 | - | - | +0.5 | 1024 | +2.5M | +2.5M |
| Paramanu-Hindi | 367M | 1 | - | 24.20 | 30.97 | - | - | 239 | 1024 | 66B | 66B |
| Paramanu-Hindi-instruct | 367M | 1 | - | 25.54 | **40.14** | - | - | +1 | 1024 | +13M | +13M |
| Paramanu-Tamil | 208M | 1 | - | - | - | 33.34 | - | 112.5 | 1024 | 39.32B | 39.32B |
| Paramanu-Tamil-instruct | 208M | 1 | - | - | - | **34.80** | - | +0.5 | 1024 | +3M | +3M |
| Paramanu-Telugu | 208M | 1 | - | - | - | - | 32.22 | 112.5 | 1024 | 39.32B | 39.32B |
| Paramanu-Telugu-instruct | 208M | 1 | - | - | - | - | **34.50** | +0.5 | 1024 | +2.8M | +2.8M |
| BanglaT5 | 247M | 1 | 21.92 | - | - | - | - | 3,000 | 512 | 196.6B | 196.6B |
| BanglaByT5 | 300M | 1 | 21.20 | - | - | - | - | 600 | 512 | 49.15B | 49.15B |
| Bloom | 560M | 45 | 23.83 | 24.76 | 32.84 | 33.13 | 31.78 | N/A | 2048 | 350B | 7.7B |
| Bloomz (instruction-tuned) | 560M | 45 | 24.01 | 25.29 | 32.31 | 33.30 | 32.25 | N/A | 2048 | +13B | +1.4B |
| xGLM | 564M | 30 | 21.54 | 22.31 | 30.69 | 30.57 | 30.35 | 129,024 | 2048 | 500B | 20B |
| Llama-3.2 | 1B | 34 | 24.90 | 26.06 | 34.54 | 32.81 | 32.13 | 370,000 | 128,000 | 9T | N/A |
| Bloom | 1.1B | 45 | 24.75 | 25.78 | 33.87 | 32.94 | 33.04 | N/A | 2048 | 350B | 7.7B |
| Bloomz (instruction-tuned) | 1.1B | 45 | 23.45 | 23.54 | 31.87 | 32.19 | 30.67 | N/A | 2048 | +13B | +1.4B |
| mGPT | 1.3B | 61 | 22.86 | 23.30 | 31.95 | 29.73 | 30.96 | 86,016 | 2048 | 400B | 15B |
| xGLM | 1.7B | 30 | 22.24 | 21.69 | 32.09 | 30.56 | 31.12 | 129,024 | 2048 | 500B | 20B |
| Sarvam | 2B | 11 | 25.22 | 26.08 | 36.27 | **35.25** | **34.55** | 122,880 | 8192 | 4T | 2T |
| Indic-Gemma-Navrasa | 2B | 16 | 25.27 | 26.02 | 34.40 | 33.73 | 33.55 | +45 | 8192 | $\geq 2T$ | N/A |
| xGLM | 2.9B | 30 | 22.27 | 21.73 | 33.18 | 30.70 | 33.51 | 129,024 | 2048 | 500B | 20B |
| Llama-3.2 | 3B | 34 | **30.17** | **31.36** | 40.06 | **36.59** | **34.86** | 460,000 | 128,000 | 9T | N/A |
| Nemotron-Hindi | 4B | 15 | - | **29.42** | **45.33** | - | - | N/A | 4096 | 8.5T | 491.2B |
| xGLM | 4.5B | 30 | 22.88 | 24.36 | 32.83 | 30.77 | 31.24 | 129,024 | 2048 | 500B | 20B |
| Bloom | 7B | 45 | 25.47 | 25.61 | 35.92 | 33.95 | 33.28 | N/A | 2048 | 350B | 7.7B |
| Bloomz (instruction-tuned) | 7B | 45 | **37.77** | **37.68** | **40.80** | **41.41** | **39.71** | N/A | 2048 | +13B | +1.4B |
| OpenHathi (CPTLLama-2) | 7B | 28 | - | 25.40 | 35.41 | - | - | N/A | 4096 | $\geq 2T$ | $\geq 7B$ |
| Airavata (instruction-tuned) | 7B | 28 | - | 26.64 | 36.93 | - | - | N/A | 4096 | $\geq 2T$ | N/A |
| xGLM | 7.5B | 30 | 22.99 | 22.47 | 34.16 | 30.73 | 31.73 | 129,024 | 2048 | 500B | 20B |
| Llama-3 | 8B | 34 | **33.54** | **33.10** | **43.45** | **39.09** | **38.59** | 1,300,000 | 8192 | $\geq 15T$ | N/A |
| Aya23 | 8B | 23 | 25.76 | 28.69 | **43.98** | 34.12 | 31.02 | N/A | 8192 | N/A | N/A |

Table 3: Summary of Zero-Shot Benchmark Average Scores across Scripts (Bangla, Devanagari, Tamil, Telugu) and Languages. Models that performed better than our models are **underlined and bold**; the best performance of our model is in **bold**; '-' represents that monolingual and multilingual models are not evaluated on languages of different scripts which were not part of training but languages of same script were evaluated even if it was not part of training; '+' denotes additional tokens/training hours on top of pretrained models for instruction-tuning.

### 4.2 Evaluation

We evaluate our models on perplexity and downstream tasks including QA, NLI, and commonsense reasoning, across five Indian languages. We also performed human evaluation for Bangla and Hindi as discussed in Appendix C.3. Comparisons are made with 21 multilingual and Indian-adapted LLMs (200M–8B params) such as BanglaT5, BanglaByT5, Bloomz, xGLM, LLaMA-3/3.2, mGPT, Sarvam 2B, Aya23, and fine-tuned models such as OpenHathi, Nemotron-Hindi (Joshi et al., 2025), Airavata (Gala et al., 2024), and Indic-Gemma-Navrasa (Telugu-LLM-Labs, 2025). Models are grouped by size.

**Benchmarks.** We use MMLU (Hendrycks et al., 2021), ARC (Clark et al., 2018), and Belebele (Bandarkar et al., 2024) for all languages, and language-specific datasets: HellaSwag (Hindi), XNLI (Conneau et al., 2018), XStoryCloze (Lin et al., 2022) (Hindi, Telugu), and XCOPA (Ponti et al., 2020) (Tamil) as benchmarks. Evaluation uses datasets from Lai et al. (2023b) and LM Evaluation Harness (LMEval) (Sutawika et al., 2024), with limited HellaSwag support due to tool issues.

### 4.3 Results

Table 3 summarizes average performance across benchmark tasks in five Indian languages (detailed results are given in Appendix Tables 9, 11, 10, 12, 13). The instruction-tuned Paramanu models outperform, on each language, the 13 LLMs with less than 3B parameters, with the exception of Sarvam 2B which slightly outperforms the Tamil and Telugu Paramanu models; it is however trained on 4T tokens, 50× more than the Paramanu models. The Paramanu models furthermore outperform the larger models OpenHathi-7B and Airavata-7B adapted on Marathi and Hindi through continual pretraining, as well as 6 out of 10 larger (above 3B parameters) language models, including LLMs using up-sampled low-resource data as xGLM which is trained with 30 languages. Although LLaMA-3-8B and Bloomz-7B lead overall due to large-scale instruction tuning on xP3, our models remain competitive with all other models.

| Language | Configuration | MMLU | ARC | Belebele | XCOPA | XStoryCloze | HellaSwag | XNLI | Average |
|---|---|---|---|---|---|---|---|---|---|
| Bangla | full (ours) | **24.82** | **25.75** | **25.11** | - | - | - | - | **25.22** |
| | full w/o Unigram | 22.66 | 23.61 | 23.67 | - | - | - | - | 23.31 |
| | full w/o cleaning | 23.67 | 24.21 | 25.00 | - | - | - | - | 24.29 |
| | full w/o Unigram w/o cleaning | 20.55 | 21.56 | 23.33 | - | - | - | - | 21.81 |
| Hindi | full (ours) | **25.18** | **27.14** | **26.22** | - | **48.78** | **25.02** | **33.49** | **30.97** |
| | full w/o Unigram | 23.35 | 25.54 | 25.22 | - | 46.78 | 24.89 | **33.49** | 29.87 |
| | full w/o cleaning | 24.72 | 22.25 | 25.44 | - | 48.16 | 24.06 | 32.34 | 29.49 |
| | full w/o Unigram w/o cleaning | 22.75 | 21.83 | 23.56 | - | 44.96 | 24.02 | 33.21 | 28.38 |
| Marathi | full (ours) | **25.39** | **26.49** | **27.33** | - | - | - | - | **26.40** |
| | full w/o Unigram | 25.31 | 23.20 | 26.11 | - | - | - | - | 24.87 |
| | full w/o cleaning | 22.47 | 22.42 | 24.00 | - | - | - | - | 22.96 |
| | full w/o Unigram w/o cleaning | 23.17 | 21.82 | 21.78 | - | - | - | - | 22.26 |
| Tamil | full (ours) | **24.37** | **24.51** | **26.88** | **57.60** | - | - | - | **33.34** |
| | full w/o Unigram | 23.87 | 24.08 | 26.11 | 57.00 | - | - | - | 32.76 |
| | full w/o cleaning | 23.25 | 23.38 | 26.56 | 56.20 | - | - | - | 32.35 |
| | full w/o Unigram w/o cleaning | 22.47 | 22.50 | 24.00 | 54.80 | - | - | - | 30.94 |
| Telugu | full (ours) | **25.26** | **26.32** | **26.00** | - | **54.20** | - | - | **32.95** |
| | full w/o Unigram | 25.12 | 21.75 | 24.44 | - | 52.42 | - | - | 30.93 |
| | full w/o cleaning | 24.16 | 22.68 | 23.11 | - | 53.73 | - | - | 30.92 |
| | full w/o Unigram w/o cleaning | 22.95 | 17.81 | 23.22 | - | 51.72 | - | - | 28.92 |

Table 4: Ablation study across Bangla (108M), Marathi (208M), Tamil (208M), Telugu (208M), and Hindi (367M) models. Evaluates the impact of tokenizer type and data cleaning. All scores are reported as zero-shot Accuracy (%). Dash (-) indicates benchmark not applicable or not available on LM-Eval.

### 4.4 Ablation Studies

**Impact of tokenizer and data cleaning.** Table 4 presents the impact of incorporating Unigram tokens into BPE and of data cleaning across five languages. Our tokenizer improves downstream task average performance over standard BPE by 1.91% (Bangla), 1.41% (Hindi), 1.53% (Marathi), 0.56% (Tamil), and 2.02% (Telugu). Additional data cleaning further boosts scores by 1% (Bangla), 3.5% (Marathi), 1% (Tamil), and 2% (Telugu), with a similar trend observed for BPE tokenizers underscoring the value of data preprocessing irrespective of tokenization strategy.

**Impact of shrinking factor for position interpolation context window.** Table 5 shows the impact of $\alpha$ for position interpolation or downscale the position token indices to accommodate more tokens than max context window possible on limited memory. The average benchmark score monotonically increased with bigger context size for all languages except Telugu. Although we gain on larger sequences, on the flip side, interpolating token position indices to reside in a much narrower region might inject noise which may perturb language modeling benchmark performance.

## 5 Discussion

### 5.1 Scaling

In our study, smaller Hindi models (e.g., 162M) often outperformed larger ones (e.g., 367M) at fixed training durations, as shown in Appendix Tables 10, 11. With extended training, larger models (e.g., 367M) surpassed smaller counterparts (e.g., 162M), underscoring the need for size-appropriate training duration. Hence, we confirm prior findings (Hoffmann et al., 2022b; Kaplan et al., 2020) that larger models require proportionally more training to outperform smaller ones.

### 5.2 Instruction-tuning Gains

*Instruction tuning improves downstream performance across all five evaluated languages.* Hindi shows the largest improvement (+9%), followed by Bangla (+4.3%). Gains are smaller for Tamil (+1.46%) and Telugu (+2.28%), likely due to lower-quality machine translation artifacts in instruction tuning datasets.

### 5.3 Cross-lingual Transfer

As one can note form Table 3, the 208M Marathi model matches the 367M Hindi model on Hindi benchmarks (30.97) due to lower perplexity (8.94 vs 11.05), indicating effective cross-lingual transfer between Hindi and Marathi that share the Devanagari script. Instruction tuning however shows asymmetry: Marathi to Hindi reduces performance ($-0.43$), while Hindi to Marathi improves it ($+1.34$), suggesting reverse transfer from high (Hindi) to low-resource (Marathi) languages (details in Table 6). Thus, these monolingual models are well suited for cross-lingual downstream tasks such as translation and information retrieval.

To further test how languages of Devanagari script generalize to unseen languages, we trained Sanskrit monolingual model and a multilingual model, mParamanu-162M on languages of Indo-European family (Assamese, Bangla, Devanagari script (Sanskrit, Konkani, Maithili), Odia). We intentionally kept out Hindi and Marathi from the training of our multilingual model to test its generalization to languages of Devanagari script. Table 6 shows strong zero-shot transfer within languages that share the same script, Devanagari.

| Language | Configuration | MMLU | ARC | Belebele | XCOPA | XStoryCloze | HellaSwag | XNLI | Average |
|---|---|---|---|---|---|---|---|---|---|
| Bangla | $\alpha$ =1, ctx=256 | 22.62 | 24.04 | 23.00 | - | - | - | - | 23.22 |
| | $\alpha$ =2, ctx=512 | **25.84** | 23.52 | **25.44** | - | - | - | - | 24.93 |
| | $\alpha$ =3, ctx=768 | 23.39 | **25.75** | 25.33 | - | - | - | - | 24.82 |
| | $\alpha$ =4, ctx=1024 | 24.82 | **25.75** | 25.22 | - | - | - | - | **25.22** |
| Hindi | $\alpha$ =2, ctx=512 | 25.43 | 25.34 | 27.56 | - | 48.78 | 25.14 | 33.17 | 30.90 |
| | $\alpha$ =3, ctx=768 | 25.36 | 25.86 | 26.44 | - | 47.78 | 25.13 | 33.29 | 30.64 |
| | $\alpha$ =4, ctx=1024 | 25.18 | 27.14 | 26.22 | - | 48.78 | 25.02 | 33.49 | 30.97 |
| Marathi | $\alpha$ =1, ctx=256 | 22.96 | **28.48** | 22.33 | - | - | - | - | 24.59 |
| | $\alpha$ =2, ctx=512 | 24.45 | 26.06 | 25.89 | - | - | - | - | 25.46 |
| | $\alpha$ =3, ctx=768 | 24.58 | 24.66 | **27.78** | - | - | - | - | 25.67 |
| | $\alpha$ =4, ctx=1024 | **25.39** | 26.94 | 27.33 | - | - | - | - | **26.55** |
| Tamil | $\alpha$ =1, ctx=256 | 22.72 | **25.04** | 23.22 | 53.20 | - | - | - | 31.04 |
| | $\alpha$ =2, ctx=512 | 23.40 | 22.94 | 23.89 | 53.80 | - | - | - | 31.00 |
| | $\alpha$ =3, ctx=768 | **25.30** | 23.91 | 23.33 | 54.80 | - | - | - | 31.84 |
| | $\alpha$ =4, ctx=1024 | 24.37 | 24.51 | **26.88** | **57.60** | - | - | - | **33.34** |
| Telugu | $\alpha$ =1, ctx=256 | **26.83** | 25.26 | 21.89 | - | 53.01 | - | - | 31.75 |
| | $\alpha$ =2, ctx=512 | 25.06 | **25.61** | **28.89** | - | 53.34 | - | - | **33.22** |
| | $\alpha$ =3, ctx=768 | 25.40 | 26.32 | 22.67 | - | 53.08 | - | - | 31.87 |
| | $\alpha$ =4, ctx=1024 | 25.26 | 26.32 | 26.00 | - | **54.20** | - | - | 32.95 |

Table 5: Ablation study for shrinking factor $\alpha$ of position interpolation for varying context size (ctx) pretraining across Bangla (108M), Marathi (208M), Tamil (208M), Telugu (208M), and Hindi (367M) models for . All scores are reported as zero-shot Accuracy (%). Dash(-) indicates benchmark not applicable or not available on LM-Eval.

| Model | Size | #langs | Devanagari | |
|---|---|---|---|---|
| | | | Marathi | Hindi |
| Paramanu-Sanskrit | 139M | 1 | 25.26 | 31.05 |
| mParamanu | 162M | 6 | 25.28 | 30.07 |
| Paramanu-Marathi | 208M | 1 | 26.40 | 30.97 |
| Paramanu-Marathi-instruct | 208M | 1 | 26.93 | 30.54 |
| Paramanu-Hindi | 367M | 1 | 24.20 | 30.97 |
| Paramanu-Hindi-instruct | 367M | 1 | 25.54 | 40.14 |

Table 6: Average zero-shot benchmark scores for cross-lingual transfer among Devanagari languages (Hindi, Marathi, Sanskrit). Sanskrit and mParamanu were not trained on Hindi.

For example, Paramanu-Sanskrit 139M, trained without Hindi, achieves 31.05 avg on Hindi, outperforming xGLM 564M and approaching Bloom 560M. mParamanu-162M also generalizes well to Hindi and Marathi. This implies that both monolingual and multilingual small language models of shared script can be further used for downstream cross-lingual NLP, information retrieval and translation tasks.

### 5.4 N-shot Degradation

The experiments and results reported in Appendix C.2 show that performance drops from 0-shot to 25-shot on XNLI-Hindi, XStoryCloze, and XCOPA, echoing trends in GPT models (Valmeekam et al., 2024). This may result from in-context examples acting as soft constraints that hinder generation.

## 6 Conclusion

We introduce *Paramanu*, the first family of open-source, Indian-only sub-400M decoder language models trained from scratch on open-source, fully transparent, language-specific data, and designed to be broadly usable by NLP researchers. We show that, for low-resource, morphologically rich languages, small language-specific models (under 400M parameters and trained on fewer than 70B tokens) can outperform larger alternatives, except for multilingual models exceeding 3B parameters and trained on over 1T tokens. *Thus, under constraints on model and data size, the optimal strategy is to build language-specific models with low-fertility, morphologically aligned tokenizers trained on cleaned data, rather than maximizing scale.* Our open-source models, tokenizers, and instruction-tuning datasets advance understanding of smaller LMs and provide a practical foundation for under-resourced researchers, and enable further research on Indian languages.

In the future, we aim to scale our methodology to all 22 official Indian languages, as well as other morphologically rich and agglutinative languages.

## Limitations

Building generative language models for Indian languages involves several challenges. Each stage such as data collection, pretraining, instruction tuning, and evaluation has its own limitations. Additionally, the societal and cultural implications of deploying AI in underrepresented linguistic communities are complex and beyond the full scope of this section.

**Data.** Our models are trained on a limited corpus consisting mainly of news articles, Wikipedia, and other structured sources in Indian languages, totaling only a few billion tokens. The dataset lacks diversity across key domains such as law, science, education, and general world knowledge. As a result, the models perform sub-optimally on benchmarks like MMLU, which test broad academic and professional understanding. As with any LLM, the data it is trained on determines the range and quality of its capabilities.

**Training.** Due to resource constraints, models were trained over multiple epochs on repeated tokens. Although this helps to reinforce learning in low-resource settings, it increases the risk of overfitting and can reduce generalization. Furthermore, our models were pretrained using a single A100 GPU, which significantly restricted training duration, batch size, and overall scale. As a result, the models may not have reached full convergence. The lack of compute and open-source infrastructure for Indian language models continues to be a major bottleneck. Detailed training logs are omitted due to space limitations.

**Instruction Tuning and Safety.** We instruction-tuned the models using 15,000 machine-translated instructions generated via Google Translate. This may introduce grammatical errors or semantic inconsistencies, as machine translation quality for Indic languages remains below human-level accuracy. No safety mechanisms such as prompt filtering, fact-checking, or toxicity detection were applied. The outputs shown in this paper are raw model responses, without post-processing. As a result, models may generate factually incorrect, biased, or inappropriate content. We emphasize the need for responsible use and future work on safety and alignment.

**Evaluation.** Evaluation was conducted in a zero-shot setting without task-specific fine-tuning. This limits performance, especially on complex or domain-specific tasks. Benchmarks like MMLU highlight the impact of limited training data and the absence of instruction tuning aligned with task objectives. Future improvements can be expected through more diverse data and supervised fine-tuning.

## Ethical Considerations

In this work, we advocate for greater openness in developing generative language models, especially for low-resource morphologically rich Indian languages serving more than 1 billion speakers. Open access is crucial for deepening our scientific understanding of these models and ensuring that communities beyond the Global North can actively participate in their advancement. Training on openly available datasets not only supports transparency but also helps bridge the gap for languages and regions that have historically been underrepresented in AI research.

By releasing our models and tokenizers openly, we empower researchers, developers, and communities to build upon existing work rather than starting from scratch saving resources and reducing environmental impact. While we acknowledge the risks of misuse, we believe that broader access enables more diverse efforts to identify, study, and mitigate potential harms. We recognize that openness comes with risks these models could be misused. However, we believe open access also helps researchers identify and reduce such risks more effectively by encouraging diverse solutions.

## References

Julien Abadji, Pedro Ortiz Suarez, Laurent Romary, and Benoît Sagot. 2022. Towards a cleaner document-oriented multilingual crawled corpus.

Kabir Ahuja, Harshita Diddee, Rishav Hada, Millicent Ochieng, Krithika Ramesh, Prachi Jain, Akshay Nambi, Tanuja Ganu, Sameer Segal, Mohamed Ahmed, Kalika Bali, and Sunayana Sitaram. 2023. MEGA: Multilingual evaluation of generative AI. In *Proceedings of the 2023 Conference on Empirical Methods in Natural Language Processing*, pages 4232–4267, Singapore. Association for Computational Linguistics.

Ebtesam Almazrouei, Hamza Alobeidli, Abdulaziz Alshamsi, Alessandro Cappelli, Ruxandra Cojocaru, Mérouane Debbah, Étienne Goffinet, Daniel Hesslow, Julien Launay, Quentin Malartic, Daniele Mazzotta, Badreddine Noune, Baptiste Pannier, and Guilherme Penedo. 2023. The falcon series of open language models.

Duarte Miguel Alves, José Pombal, Nuno M Guerreiro, Pedro Henrique Martins, João Alves, Amin Farajian, Ben Peters, Ricardo Rei, Patrick Fernandes, Sweta Agrawal, Pierre Colombo, José G. C. de Souza, and Andre Martins. 2024. Tower: An open multilingual large language model for translation-related tasks. In *First Conference on Language Modeling*.

Viraat Aryabumi, John Dang, Dwarak Talupuru, Saurabh Dash, David Cairuz, Hangyu Lin, Bharat Venkitesh, Madeline Smith, Jon Ander Campos, Yi Chern Tan, Kelly Marchisio, Max Bartolo, Sebastian Ruder, Acyr Locatelli, Julia Kreutzer, Nick Frosst, Aidan Gomez, Phil Blunsom, Marzieh Fadaee, Ahmet Üstün, and Sara Hooker. 2024. Aya 23: Open weight releases to further multilingual progress.

Lucas Bandarkar, Davis Liang, Benjamin Muller, Mikel Artetxe, Satya Narayan Shukla, Donald Husa, Naman Goyal, Abhinandan Krishnan, Luke Zettlemoyer, and Madian Khabsa. 2024. The belebele benchmark: a parallel reading comprehension dataset in 122 language variants. In *Proceedings of the 62nd Annual Meeting of the Association for Computational Linguistics (Volume 1: Long Papers)*, pages 749–775, Bangkok, Thailand. Association for Computational Linguistics.

Yejin Bang, Samuel Cahyawijaya, Nayeon Lee, Wenliang Dai, Dan Su, Bryan Wilie, Holy Lovenia, Ziwei Ji, Tiezheng Yu, Willy Chung, Quyet V. Do, Yan Xu, and Pascale Fung. 2023. A multitask, multilingual, multimodal evaluation of ChatGPT on reasoning, hallucination, and interactivity. In *Proceedings of the 13th International Joint Conference on Natural Language Processing and the 3rd Conference of the Asia-Pacific Chapter of the Association for Computational Linguistics (Volume 1: Long Papers)*, pages 675–718, Nusa Dua, Bali. Association for Computational Linguistics.

Scott Barnett, Zac Brannelly, Stefanus Kurniawan, and Sheng Wong. 2024. Fine-tuning or fine-failing? debunking performance myths in large language models.

Abhik Bhattacharjee, Tahmid Hasan, Wasi Uddin Ahmad, and Rifat Shahriyar. 2023. BanglaNLG and BanglaT5: Benchmarks and resources for evaluating low-resource natural language generation in Bangla. In *Findings of the Association for Computational Linguistics: EACL 2023*, pages 726–735, Dubrovnik, Croatia. Association for Computational Linguistics.

Pramit Bhattacharyya and Arnab Bhattacharya. 2025. BanglaByT5: Byte-level modelling for Bangla. In *Findings of the Association for Computational Linguistics: EMNLP 2025*, pages 5551–5560, Suzhou, China. Association for Computational Linguistics.

Pramit Bhattacharyya, Joydeep Mondal, Subhadip Maji, and Arnab Bhattacharya. 2023. VACASPATI: A diverse corpus of Bangla literature. In *Proceedings of the 13th International Joint Conference on Natural Language Processing and the 3rd Conference of the Asia-Pacific Chapter of the Association for Computational Linguistics (Volume 1: Long Papers)*, pages 1118–1130, Nusa Dua, Bali. Association for Computational Linguistics.

Sid Black, Stella Biderman, Eric Hallahan, Quentin Anthony, Leo Gao, Laurence Golding, Horace He, Connor Leahy, Kyle McDonell, Jason Phang, Michael Pieler, USVSN Sai Prashanth, Shivanshu Purohit, Laria Reynolds, Jonathan Tow, Ben Wang, and Samuel Weinbach. 2022. Gpt-neox-20b: An open-source autoregressive language model.

Kaj Bostrom and Greg Durrett. 2020. Byte pair encoding is suboptimal for language model pretraining. In *Findings of the Association for Computational Linguistics: EMNLP 2020*, pages 4617–4624, Online. Association for Computational Linguistics.

Tyler A. Chang, Catherine Arnett, Zhuowen Tu, and Ben Bergen. 2024. When is multilinguality a curse? language modeling for 250 high- and low-resource languages. In *Proceedings of the 2024 Conference on Empirical Methods in Natural Language Processing*, pages 4074–4096, Miami, Florida, USA. Association for Computational Linguistics.

Shouyuan Chen, Sherman Wong, Liangjian Chen, and Yuandong Tian. 2023. Extending context window of large language models via positional interpolation.

Monojit Choudhury. 2023. Generative ai has a language problem. *Nature human behaviour*, 7(11):1802–1803.

Aakanksha Chowdhery, Sharan Narang, Jacob Devlin, Maarten Bosma, Gaurav Mishra, Adam Roberts, Paul Barham, Hyung Won Chung, Charles Sutton, Sebastian Gehrmann, Parker Schuh, Kensen Shi, Sasha Tsvyashchenko, Joshua Maynez, Abhishek

Rao, Parker Barnes, Yi Tay, Noam Shazeer, Vinodkumar Prabhakaran, Emily Reif, Nan Du, Ben Hutchinson, Reiner Pope, James Bradbury, Jacob Austin, Michael Isard, Guy Gur-Ari, Pengcheng Yin, Toju Duke, Anselm Levskaya, Sanjay Ghemawat, Sunipa Dev, Henryk Michalewski, Xavier Garcia, Vedant Misra, Kevin Robinson, Liam Fedus, Denny Zhou, Daphne Ippolito, David Luan, Hyeontaek Lim, Barret Zoph, Alexander Spiridonov, Ryan Sepassi, David Dohan, Shivani Agrawal, Mark Omernick, Andrew M. Dai, Thanumalayan Sankaranarayana Pillai, Marie Pellat, Aitor Lewkowycz, Erica Moreira, Rewon Child, Oleksandr Polozov, Katherine Lee, Zongwei Zhou, Xuezhi Wang, Brennan Saeta, Mark Diaz, Orhan Firat, Michele Catasta, Jason Wei, Kathy Meier-Hellstern, Douglas Eck, Jeff Dean, Slav Petrov, and Noah Fiedel. 2023. Palm: Scaling language modeling with pathways. *Journal of Machine Learning Research*, 24(240):1–113.

Peter Clark, Isaac Cowhey, Oren Etzioni, Tushar Khot, Ashish Sabharwal, Carissa Schoenick, and Oyvind Tafjord. 2018. Think you have solved question answering? try arc, the ai2 reasoning challenge.

Alexis Conneau, Kartikay Khandelwal, Naman Goyal, Vishrav Chaudhary, Guillaume Wenzek, Francisco Guzmán, Edouard Grave, Myle Ott, Luke Zettlemoyer, and Veselin Stoyanov. 2020. Unsupervised cross-lingual representation learning at scale. In *Proceedings of the 58th Annual Meeting of the Association for Computational Linguistics*, pages 8440–8451, Online. Association for Computational Linguistics.

Alexis Conneau, Ruty Rinott, Guillaume Lample, Adina Williams, Samuel Bowman, Holger Schwenk, and Veselin Stoyanov. 2018. XNLI: Evaluating cross-lingual sentence representations. In *Proceedings of the 2018 Conference on Empirical Methods in Natural Language Processing*, pages 2475–2485, Brussels, Belgium. Association for Computational Linguistics.

Mike Conover, Matt Hayes, Ankit Mathur, Jianwei Xie, Jun Wan, Sam Shah, Ali Ghodsi, Patrick Wendell, Matei Zaharia, and Reynold Xin. 2023. Free dolly: Introducing the world's first truly open instruction-tuned llm.

Chaitali Dangarikar, Arnab Bhattacharya, Karthika N J, Chaitanya S Lakkundi, Ganesh Ramakrishnan, Annarao Kulkarni, Shivani V, Pramit Bhattacharyya, and Hrishikesh Terdalkar. 2024. Samanvaya: An interlingua for the unity of indian languages. *Central Sanskrit University*.

Tim Dettmers, Artidoro Pagnoni, Ari Holtzman, and Luke Zettlemoyer. 2023. QLoRA: Efficient finetuning of quantized LLMs. In *Thirty-seventh Conference on Neural Information Processing Systems*.

Jacob Devlin, Ming-Wei Chang, Kenton Lee, and Kristina Toutanova. 2019. BERT: Pre-training of deep bidirectional transformers for language understanding. In *Proceedings of the 2019 Conference of the North American Chapter of the Association for Computational Linguistics: Human Language Technologies, Volume 1 (Long and Short Papers)*, pages 4171–4186, Minneapolis, Minnesota. Association for Computational Linguistics.

Sumanth Doddapaneni, Rahul Aralikatte, Gowtham Ramesh, Shreya Goyal, Mitesh M. Khapra, Anoop Kunchukuttan, and Pratyush Kumar. 2023. Towards leaving no Indic language behind: Building monolingual corpora, benchmark and models for Indic languages. In *Proceedings of the 61st Annual Meeting of the Association for Computational Linguistics (Volume 1: Long Papers)*, pages 12402–12426, Toronto, Canada. Association for Computational Linguistics.

Abhimanyu Dubey, Abhinav Jauhri, Abhinav Pandey, Abhishek Kadian, Ahmad Al-Dahle, Aiesha Letman, Akhil Mathur, Alan Schelten, Amy Yang, Angela Fan, Anirudh Goyal, Anthony Hartshorn, Aobo Yang, Archi Mitra, Archie Sravankumar, Artem Korenev, Arthur Hinsvark, Arun Rao, Aston Zhang, Aurelien Rodriguez, Austen Gregerson, Ava Spataru, Baptiste Roziere, Bethany Biron, Binh Tang, Bobbie Chern, Charlotte Caucheteux, Chaya Nayak, Chloe Bi, Chris Marra, Chris McConnell, Christian Keller, Christophe Touret, Chunyang Wu, Corinne Wong, Cristian Canton Ferrer, Cyrus Nikolaidis, Damien Allonsius, Daniel Song, Danielle Pintz, Danny Livshits, David Esiobu, Dhruv Choudhary, Dhruv Mahajan, Diego Garcia-Olano, Diego Perino, Dieuwke Hupkes, Egor Lakomkin, Ehab AlBadawy, Elina Lobanova, Emily Dinan, Eric Michael Smith, Filip Radenovic, Frank Zhang, Gabriel Synnaeve, Gabrielle Lee, Georgia Lewis Anderson, Graeme Nail, Gregoire Mialon, Guan Pang, Guillem Cucurell, Hailey Nguyen, Hannah Korevaar, Hu Xu, Hugo Touvron, Iliyan Zarov, Imanol Arrieta Ibarra, Isabel Kloumann, Ishan Misra, Ivan Evtimov, Jade Copet, Jaewon Lee, Jan Geffert, Jana Vranes, Jason Park, Jay Mahadeokar, Jeet Shah, Jelmer van der Linde, Jennifer Billock, Jenny Hong, Jenya Lee, Jeremy Fu, Jianfeng Chi, Jianyu Huang, Jiawen Liu, Jie Wang, Jiecao Yu, Joanna Bitton, Joe Spisak, Jongsoo Park, Joseph Rocca, Joshua Johnstun, Joshua Saxe, Junteng Jia, Kalyan Vasuden Alwala, Kartikeya Upasani, Kate Plawiak, Ke Li, Kenneth Heafield, Kevin Stone, Khalid El-Arini, Krithika Iyer, Kshitiz Malik, Kuenley Chiu, Kunal Bhalla, Lauren Rantala-Yeary, Laurens van der Maaten, Lawrence Chen, Liang Tan, Liz Jenkins, Louis Martin, Lovish Madaan, Lubo Malo, Lukas Blecher, Lukas Landzaat, Luke de Oliveira, Madeline Muzzi, Mahesh Pasupuleti, Mannat Singh, Manohar Paluri, Marcin Kardas, Mathew Oldham, Mathieu Rita, Maya Pavlova, Melanie Kambadur, Mike Lewis, Min Si, Mitesh Kumar Singh, Mona Hassan, Naman Goyal, Narjes Torabi, Nikolay Bashlykov, Nikolay Bogoychev, Niladri Chatterji, Olivier Duchenne, Onur Çelebi, Patrick Alrassy, Pengchuan Zhang, Peng-wei Li, Petar Vasic, Peter Weng, Prajjwal Bhargava, Pratik Dubal, Praveen Krishnan, Punit Singh Koura, Puxin Xu, Qing He, Qingxiao Dong, Ragavan Srinivasan, Raj Ganapathy, Ramon Calderer, Ricardo Silveira Cabral, Robert Stojnic, Roberta Raileanu, Rohit Girdhar, Rohit Patel, Romain Sauvestre, Ronnie Polidoro, Roshan Sumbaly, Ross Taylor, Ruan Silva, Rui Hou, Rui Wang, Saghar Hosseini, Sahana Chennabasappa, Sanjay Singh, Sean Bell, Seohyun Sonia Kim, Sergey Edunov, Shaoliang Nie, Sharan Narang, Sharath Raparthy, Sheng Shen, Shengye Wan, Shruti Bhosale, Shun Zhang, Simon Vandenhende, Soumya Batra, Spencer Whitman, Sten Sootla, Stephane Collot, Suchin Gururangan, Sydney Borodinsky, Tamar Herman, Tara Fowler, Tarek Sheasha, Thomas Georgiou, Thomas Scialom, Tobias Speckbacher, Todor Mihaylov, Tong Xiao, Ujjwal Karn, Vedanuj Goswami, Vibhor Gupta, Vignesh Ramanathan, Viktor Kerkez, Vincent Gonguet, Virginie Do, Vish Vogeti, Vladan Petrovic, Weiwei Chu, Wenhan Xiong, Wenyin Fu, Whitney Meers, Xavier Martinet, Xiaodong Wang, Xiaoqing Ellen Tan, Xinfeng Xie, Xuchao Jia, Xuewei Wang, Yaelle Goldschlag, Yashesh Gaur, Yasmine Babaei, Yi Wen, Yiwen Song, Yuchen Zhang, Yue Li, Yuning Mao, Zacharie Delpierre Coudert, Zheng Yan, Zhengxing Chen, Zoe Papakipos, Aaditya Singh, Aaron Grattafiori, Abha Jain, Adam Kelsey, Adam Shajnfeld, Adithya Gangidi, Adolfo Victoria, Ahuva Goldstand, Ajay Menon, Ajay Sharma, Alex Boesenberg, Alex Vaughan, Alexei Baevski, Allie Feinstein, Amanda Kallet, Amit Sangani, Anam Yunus, Andrei Lupu, Andres Alvarado, Andrew Caples, Andrew Gu, Andrew Ho, Andrew Poulton, Andrew Ryan, Ankit Ramchandani, Annie Franco, Aparajita Saraf, Arkabandhu Chowdhury, Ashley Gabriel, Ashwin Bharambe, Assaf Eisenman, Azadeh Yazdan, Beau James, Ben Maurer, Benjamin Leonhardi, Bernie Huang, Beth Loyd, Beto De Paola, Bhargavi Paranjape, Bing Liu, Bo Wu, Boyu Ni, Braden Hancock, Bram Wasti, Brandon Spence, Brani Stojkovic, Brian Gamido, Britt Montalvo, Carl Parker, Carly Burton, Catalina Mejia, Changhan Wang, Changkyu Kim, Chao Zhou, Chester Hu, Ching-Hsiang Chu, Chris Cai, Chris Tindal, Christoph Feichtenhofer, Damon Civin, Dana Beaty, Daniel Kreymer, Daniel Li, Danny Wyatt, David Adkins, David Xu, Davide Testuggine, Delia David, Devi Parikh, Diana Liskovich, Didem Foss, Dingkang Wang, Duc Le, Dustin Holland, Edward Dowling, Eissa Jamil, Elaine Montgomery, Eleonora Presani, Emily Hahn, Emily Wood, Erik Brinkman, Esteban Arcaute, Evan Dunbar, Evan Smothers, Fei Sun, Felix Kreuk, Feng Tian, Firat Ozgenel, Francesco Caggioni, Francisco Guzmán, Frank Kanayet, Frank Seide, Gabriela Medina Florez, Gabriella Schwarz, Gada Badeer, Georgia Swee, Gil Halpern, Govind Thattai, Grant Herman, Grigory Sizov, Guangyi, Zhang, Guna Lakshminarayanan, Hamid Shojanazeri, Han Zou, Hannah Wang, Hanwen Zha, Haroun Habeeb, Harrison Rudolph, Helen Suk, Henry Aspegren, Hunter Goldman, Ibrahim Damlaj, Igor Molybog, Igor Tufanov, Irina-Elena Veliche, Itai Gat, Jake Weissman, James Geboski, James Kohli, Japhet Asher, Jean-Baptiste Gaya, Jeff Marcus, Jeff Tang, Jennifer Chan, Jenny Zhen, Jeremy Reizenstein, Jeremy Teboul, Jessica Zhong, Jian Jin, Jingyi Yang, Joe Cummings, Jon Carvill, Jon Shepard, Jonathan McPhie, Jonathan Torres, Josh Ginsburg, Junjie Wang, Kai Wu, Kam Hou U, Karan Saxena, Karthik Prasad, Kartikay Khandelwal, Katayoun Zand, Kathy Matosich, Kaushik Veeraraghavan, Kelly Michelena, Keqian Li, Kun Huang, Kunal Chawla, Kushal Lakhotia, Kyle Huang, Lailin Chen, Lakshya Garg, Lavender A, Leandro Silva, Lee Bell, Lei Zhang, Liangpeng Guo, Licheng Yu, Liron Moshkovich, Luca Wehrstedt, Madian Khabsa, Manav Avalani, Manish Bhatt, Maria Tsimpoukelli, Martynas Mankus, Matan Hasson, Matthew Lennie, Matthias Reso, Maxim Groshev, Maxim Naumov, Maya Lathi, Meghan Keneally, Michael L. Seltzer, Michal Valko, Michelle Restrepo, Mihir Patel, Mik Vyatskov, Mikayel Samvelyan, Mike Clark, Mike Macey, Mike Wang, Miquel Jubert Hermoso, Mo Metanat, Mohammad Rastegari, Munish Bansal, Nandhini Santhanam, Natascha Parks, Natasha White, Navyata Bawa, Nayan Singhal, Nick Egebo, Nicolas Usunier, Nikolay Pavlovich Laptev, Ning Dong, Ning Zhang, Norman Cheng, Oleg Chernoguz, Olivia Hart, Omkar Salpekar, Ozlem Kalinli, Parkin Kent, Parth Parekh, Paul Saab, Pavan Balaji, Pedro Rittner, Philip Bontrager, Pierre Roux, Piotr Dollar, Polina Zvyagina, Prashant Ratanchandani, Pritish Yuvraj, Qian Liang, Rachad Alao, Rachel Rodriguez, Rafi Ayub, Raghotham Murthy, Raghu Nayani, Rahul Mitra, Raymond Li, Rebekkah Hogan, Robin Battey, Rocky Wang, Rohan Maheswari, Russ Howes, Ruty Rinott, Sai Jayesh Bondu, Samyak Datta, Sara Chugh, Sara Hunt, Sargun Dhillon, Sasha Sidorov, Satadru Pan, Saurabh Verma, Seiji Yamamoto, Sharadh Ramaswamy, Shaun Lindsay, Shaun Lindsay, Sheng Feng, Shenghao Lin, Shengxin Cindy Zha, Shiva Shankar, Shuqiang Zhang, Shuqiang Zhang, Sinong Wang, Sneha Agarwal, Soji Sajuyigbe, Soumith Chintala, Stephanie Max, Stephen Chen, Steve Kehoe, Steve Satterfield, Sudarshan Govindaprasad, Sumit Gupta, Sungmin Cho, Sunny Virk, Suraj Subramanian, Sy Choudhury, Sydney Goldman, Tal Remez, Tamar Glaser, Tamara Best, Thilo Kohler, Thomas Robinson, Tianhe Li, Tianjun Zhang, Tim Matthews, Timothy Chou, Tzook Shaked, Varun Vontimitta, Victoria Ajayi, Victoria Montanez, Vijai Mohan, Vinay Satish Kumar, Vishal Mangla, Vítor Albiero, Vlad Ionescu, Vlad Poenaru, Vlad Tiberiu Mihailescu, Vladimir Ivanov, Wei Li, Wenchen Wang, Wenwen Jiang, Wes Bouaziz, Will Constable, Xiaocheng Tang, Xiaofang Wang, Xiaojian Wu, Xiaolan Wang, Xide Xia, Xilun Wu, Xinbo Gao, Yanjun Chen, Ye Hu, Ye Jia, Ye Qi, Yenda Li, Yilin Zhang, Ying Zhang, Yossi Adi, Youngjin Nam, Yu, Wang, Yuchen Hao, Yundi Qian, Yuzi He, Zach Rait, Zachary DeVito, Zef Rosnbrick, Zhaoduo Wen, Zhenyu Yang, and Zhiwei Zhao. 2024. The llama 3 herd of models.

Jay Gala, Thanmay Jayakumar, Jaavid Aktar Hu-

sain, Aswanth Kumar M, Mohammed Safi Ur Rahman Khan, Diptesh Kanojia, Ratish Puduppully, Mitesh M. Khapra, Raj Dabre, Rudra Murthy, and Anoop Kunchukuttan. 2024. Airavata: Introducing hindi instruction-tuned llm.

Naman Goyal, Cynthia Gao, Vishrav Chaudhary, Peng-Jen Chen, Guillaume Wenzek, Da Ju, Sanjana Krishnan, MarcAurelio Ranzato, Francisco Guzmán, and Angela Fan. 2022. The flores-101 evaluation benchmark for low-resource and multilingual machine translation. *Transactions of the Association for Computational Linguistics*, 10:522–538.

Aaron Grattafiori, Abhimanyu Dubey, Abhinav Jauhri, Abhinav Pandey, Abhishek Kadian, Ahmad Al-Dahle, Aiesha Letman, Akhil Mathur, Alan Schelten, Alex Vaughan, Amy Yang, Angela Fan, Anirudh Goyal, Anthony Hartshorn, Aobo Yang, Archi Mitra, Archie Sravankumar, Artem Korenev, Arthur Hinsvark, Arun Rao, Aston Zhang, Aurelien Rodriguez, Austen Gregerson, Ava Spataru, Baptiste Roziere, Bethany Biron, Binh Tang, Bobbie Chern, Charlotte Caucheteux, Chaya Nayak, Chloe Bi, Chris Marra, Chris McConnell, Christian Keller, Christophe Touret, Chunyang Wu, Corinne Wong, Cristian Canton Ferrer, Cyrus Nikolaidis, Damien Allonsius, Daniel Song, Danielle Pintz, Danny Livshits, Danny Wyatt, David Esiobu, Dhruv Choudhary, Dhruv Mahajan, Diego Garcia-Olano, Diego Perino, Dieuwke Hupkes, Egor Lakomkin, Ehab AlBadawy, Elina Lobanova, Emily Dinan, Eric Michael Smith, Filip Radenovic, Francisco Guzmán, Frank Zhang, Gabriel Synnaeve, Gabrielle Lee, Georgia Lewis Anderson, Govind Thattai, Graeme Nail, Gregoire Mialon, Guan Pang, Guillem Cucurell, Hailey Nguyen, Hannah Korevaar, Hu Xu, Hugo Touvron, Iliyan Zarov, Imanol Arrieta Ibarra, Isabel Kloumann, Ishan Misra, Ivan Evtimov, Jack Zhang, Jade Copet, Jaewon Lee, Jan Geffert, Jana Vranes, Jason Park, Jay Mahadeokar, Jeet Shah, Jelmer van der Linde, Jennifer Billock, Jenny Hong, Jenya Lee, Jeremy Fu, Jianfeng Chi, Jianyu Huang, Jiawen Liu, Jie Wang, Jiecao Yu, Joanna Bitton, Joe Spisak, Jongsoo Park, Joseph Rocca, Joshua Johnstun, Joshua Saxe, Junteng Jia, Kalyan Vasuden Alwala, Karthik Prasad, Kartikeya Upasani, Kate Plawiak, Ke Li, Kenneth Heafield, Kevin Stone, Khalid El-Arini, Krithika Iyer, Kshitiz Malik, Kuenley Chiu, Kunal Bhalla, Kushal Lakhotia, Lauren Rantala-Yeary, Laurens van der Maaten, Lawrence Chen, Liang Tan, Liz Jenkins, Louis Martin, Lovish Madaan, Lubo Malo, Lukas Blecher, Lukas Landzaat, Luke de Oliveira, Madeline Muzzi, Mahesh Pasupuleti, Mannat Singh, Manohar Paluri, Marcin Kardas, Maria Tsimpoukelli, Mathew Oldham, Mathieu Rita, Maya Pavlova, Melanie Kambadur, Mike Lewis, Min Si, Mitesh Kumar Singh, Mona Hassan, Naman Goyal, Narjes Torabi, Nikolay Bashlykov, Nikolay Bogoychev, Niladri Chatterji, Ning Zhang, Olivier Duchenne, Onur Çelebi, Patrick Alrassy, Pengchuan Zhang, Pengwei Li, Petar Vasic, Peter Weng, Prajjwal Bhargava, Pratik Dubal, Praveen Krishnan, Punit Singh Koura, Puxin Xu, Qing He, Qingxiao Dong, Ragavan Srinivasan, Raj Ganapathy, Ramon Calderer, Ricardo Silveira Cabral, Robert Stojnic, Roberta Raileanu, Rohan Maheswari, Rohit Girdhar, Rohit Patel, Romain Sauvestre, Ronnie Polidoro, Roshan Sumbaly, Ross Taylor, Ruan Silva, Rui Hou, Rui Wang, Saghar Hosseini, Sahana Chennabasappa, Sanjay Singh, Sean Bell, Seohyun Sonia Kim, Sergey Edunov, Shaoliang Nie, Sharan Narang, Sharath Raparthy, Sheng Shen, Shengye Wan, Shruti Bhosale, Shun Zhang, Simon Vandenhende, Soumya Batra, Spencer Whitman, Sten Sootla, Stephane Collot, Suchin Gururangan, Sydney Borodinsky, Tamar Herman, Tara Fowler, Tarek Sheasha, Thomas Georgiou, Thomas Scialom, Tobias Speckbacher, Todor Mihaylov, Tong Xiao, Ujjwal Karn, Vedanuj Goswami, Vibhor Gupta, Vignesh Ramanathan, Viktor Kerkez, Vincent Gonguet, Virginie Do, Vish Vogeti, Vítor Albiero, Vladan Petrovic, Weiwei Chu, Wenhan Xiong, Wenyin Fu, Whitney Meers, Xavier Martinet, Xiaodong Wang, Xiaofang Wang, Xiaoqing Ellen Tan, Xide Xia, Xinfeng Xie, Xuchao Jia, Xuewei Wang, Yaelle Goldschlag, Yashesh Gaur, Yasmine Babaei, Yi Wen, Yiwen Song, Yuchen Zhang, Yue Li, Yuning Mao, Zacharie Delpierre Coudert, Zheng Yan, Zhengxing Chen, Zoe Papakipos, Aaditya Singh, Aayushi Srivastava, Abha Jain, Adam Kelsey, Adam Shajnfeld, Adithya Gangidi, Adolfo Victoria, Ahuva Goldstand, Ajay Menon, Ajay Sharma, Alex Boesenberg, Alexei Baevski, Allie Feinstein, Amanda Kallet, Amit Sangani, Amos Teo, Anam Yunus, Andrei Lupu, Andres Alvarado, Andrew Caples, Andrew Gu, Andrew Ho, Andrew Poulton, Andrew Ryan, Ankit Ramchandani, Annie Dong, Annie Franco, Anuj Goyal, Aparajita Saraf, Arkabandhu Chowdhury, Ashley Gabriel, Ashwin Bharambe, Assaf Eisenman, Azadeh Yazdan, Beau James, Ben Maurer, Benjamin Leonhardi, Bernie Huang, Beth Loyd, Beto De Paola, Bhargavi Paranjape, Bing Liu, Bo Wu, Boyu Ni, Braden Hancock, Bram Wasti, Brandon Spence, Brani Stojkovic, Brian Gamido, Britt Montalvo, Carl Parker, Carly Burton, Catalina Mejia, Ce Liu, Changhan Wang, Changkyu Kim, Chao Zhou, Chester Hu, Ching-Hsiang Chu, Chris Cai, Chris Tindal, Christoph Feichtenhofer, Cynthia Gao, Damon Civin, Dana Beaty, Daniel Kreymer, Daniel Li, David Adkins, David Xu, Davide Testuggine, Delia David, Devi Parikh, Diana Liskovich, Didem Foss, Dingkang Wang, Duc Le, Dustin Holland, Edward Dowling, Eissa Jamil, Elaine Montgomery, Eleonora Presani, Emily Hahn, Emily Wood, Eric-Tuan Le, Erik Brinkman, Esteban Arcaute, Evan Dunbar, Evan Smothers, Fei Sun, Felix Kreuk, Feng Tian, Filippos Kokkinos, Firat Ozgenel, Francesco Caggioni, Frank Kanayet, Frank Seide, Gabriela Medina Florez, Gabriella Schwarz, Gada Badeer, Georgia Swee, Gil Halpern, Grant Herman, Grigory Sizov, Guangyi, Zhang, Guna Lakshminarayanan, Hakan Inan, Hamid Shojanazeri, Han Zou, Hannah Wang, Hanwen Zha, Haroun Habeeb, Harrison Rudolph, Helen Suk, Henry Aspegren, Hunter Goldman, Hongyuan Zhan, Ibrahim Damlaj, Igor Molybog, Igor Tufanov, Ilias Leon-tiadis, Irina-Elena Veliche, Itai Gat, Jake Weissman, James Geboski, James Kohli, Janice Lam, Japhet Asher, Jean-Baptiste Gaya, Jeff Marcus, Jeff Tang, Jennifer Chan, Jenny Zhen, Jeremy Reizenstein, Jeremy Teboul, Jessica Zhong, Jian Jin, Jingyi Yang, Joe Cummings, Jon Carvill, Jon Shepard, Jonathan McPhie, Jonathan Torres, Josh Ginsburg, Junjie Wang, Kai Wu, Kam Hou U, Karan Saxena, Kartikay Khandelwal, Katayoun Zand, Kathy Matosich, Kaushik Veeraraghavan, Kelly Michelena, Keqian Li, Kiran Jagadeesh, Kun Huang, Kunal Chawla, Kyle Huang, Lailin Chen, Lakshya Garg, Lavender A, Leandro Silva, Lee Bell, Lei Zhang, Liangpeng Guo, Licheng Yu, Liron Moshkovich, Luca Wehrstedt, Madian Khabsa, Manav Avalani, Manish Bhatt, Martynas Mankus, Matan Hasson, Matthew Lennie, Matthias Reso, Maxim Groshev, Maxim Naumov, Maya Lathi, Meghan Keneally, Miao Liu, Michael L. Seltzer, Michal Valko, Michelle Restrepo, Mihir Patel, Mik Vyatskov, Mikayel Samvelyan, Mike Clark, Mike Macey, Mike Wang, Miquel Jubert Hermoso, Mo Metanat, Mohammad Rastegari, Munish Bansal, Nandhini Santhanam, Natascha Parks, Natasha White, Navyata Bawa, Nayan Singhal, Nick Egebo, Nicolas Usunier, Nikhil Mehta, Nikolay Pavlovich Laptev, Ning Dong, Norman Cheng, Oleg Chernoguz, Olivia Hart, Omkar Salpekar, Ozlem Kalinli, Parkin Kent, Parth Parekh, Paul Saab, Pavan Balaji, Pedro Rittner, Philip Bontrager, Pierre Roux, Piotr Dollar, Polina Zvyagina, Prashant Ratanchandani, Pritish Yuvraj, Qian Liang, Rachad Alao, Rachel Rodriguez, Rafi Ayub, Raghotham Murthy, Raghu Nayani, Rahul Mitra, Rangaprabhu Parthasarathy, Raymond Li, Rebekkah Hogan, Robin Battey, Rocky Wang, Russ Howes, Ruty Rinott, Sachin Mehta, Sachin Siby, Sai Jayesh Bondu, Samyak Datta, Sara Chugh, Sara Hunt, Sargun Dhillon, Sasha Sidorov, Satadru Pan, Saurabh Mahajan, Saurabh Verma, Seiji Yamamoto, Sharadh Ramaswamy, Shaun Lindsay, Shaun Lindsay, Sheng Feng, Shenghao Lin, Shengxin Cindy Zha, Shishir Patil, Shiva Shankar, Shuqiang Zhang, Shuqiang Zhang, Sinong Wang, Sneha Agarwal, Soji Sajuyigbe, Soumith Chintala, Stephanie Max, Stephen Chen, Steve Kehoe, Steve Satterfield, Sudarshan Govindaprasad, Sumit Gupta, Summer Deng, Sungmin Cho, Sunny Virk, Suraj Subramanian, Sy Choudhury, Sydney Goldman, Tal Remez, Tamar Glaser, Tamara Best, Thilo Koehler, Thomas Robinson, Tianhe Li, Tianjun Zhang, Tim Matthews, Timothy Chou, Tzook Shaked, Varun Vontimitta, Victoria Ajayi, Victoria Montanez, Vijai Mohan, Vinay Satish Kumar, Vishal Mangla, Vlad Ionescu, Vlad Poenaru, Vlad Tiberiu Mihailescu, Vladimir Ivanov, Wei Li, Wenchen Wang, Wenwen Jiang, Wes Bouaziz, Will Constable, Xiaocheng Tang, Xiaojian Wu, Xiaolan Wang, Xilun Wu, Xinbo Gao, Yaniv Kleinman, Yanjun Chen, Ye Hu, Ye Jia, Ye Qi, Yenda Li, Yilin Zhang, Ying Zhang, Yossi Adi, Youngjin Nam, Yu, Wang, Yu Zhao, Yuchen Hao, Yundi Qian, Yunlu Li, Yuzi He, Zach Rait, Zachary DeVito, Zef Rosnbrick, Zhaoduo Wen, Zhenyu Yang, Zhiwei Zhao, and Zhiyu Ma. 2024. The llama 3 herd of models.

Yanzhu Guo, Simone Conia, Zelin Zhou, Min Li, Saloni Potdar, and Henry Xiao. 2024a. Do large language models have an english accent? evaluating and improving the naturalness of multilingual llms.

Yanzhu Guo, Guokan Shang, and Chloé Clavel. 2024b. Benchmarking linguistic diversity of large language models. *ArXiv*, abs/2412.10271.

Dan Hendrycks, Collin Burns, Steven Basart, Andy Zou, Mantas Mazeika, Dawn Song, and Jacob Steinhardt. 2021. Measuring massive multitask language understanding. In *International Conference on Learning Representations*.

Jordan Hoffmann, Sebastian Borgeaud, Arthur Mensch, Elena Buchatskaya, Trevor Cai, Eliza Rutherford, Diego de Las Casas, Lisa Anne Hendricks, Johannes Welbl, Aidan Clark, Thomas Hennigan, Eric Noland, Katherine Millican, George van den Driessche, Bogdan Damoc, Aurelia Guy, Simon Osindero, Karén Simonyan, Erich Elsen, Oriol Vinyals, Jack Rae, and Laurent Sifre. 2022a. An empirical analysis of compute-optimal large language model training. In *Advances in Neural Information Processing Systems*, volume 35, pages 30016–30030. Curran Associates, Inc.

Jordan Hoffmann, Sebastian Borgeaud, Arthur Mensch, Elena Buchatskaya, Trevor Cai, Eliza Rutherford, Diego de Las Casas, Lisa Anne Hendricks, Johannes Welbl, Aidan Clark, Tom Hennigan, Eric Noland, Katie Millican, George van den Driessche, Bogdan Damoc, Aurelia Guy, Simon Osindero, Karen Simonyan, Erich Elsen, Jack W. Rae, Oriol Vinyals, and Laurent Sifre. 2022b. Training compute-optimal large language models.

Edward J Hu, yelong shen, Phillip Wallis, Zeyuan Allen-Zhu, Yuanzhi Li, Shean Wang, Lu Wang, and Weizhu Chen. 2022. LoRA: Low-rank adaptation of large language models. In *International Conference on Learning Representations*.

Haoyang Huang, Tianyi Tang, Dongdong Zhang, Xin Zhao, Ting Song, Yan Xia, and Furu Wei. 2023. Not all languages are created equal in LLMs: Improving multilingual capability by cross-lingual-thought prompting. In *The 2023 Conference on Empirical Methods in Natural Language Processing*.

Shaoxiong Ji, Zihao Li, Indraneil Paul, Jaakko Paavola, Peiqin Lin, Pinzhen Chen, Dayyán O'Brien, Hengyu Luo, Hinrich Schuetze, Jörg Tiedemann, and Barry Haddow. 2025. EMMA-500: Enhancing massively multilingual adaptation of large language models.

Pratik Joshi, Sebastin Santy, Amar Budhiraja, Kalika Bali, and Monojit Choudhury. 2020. The state and fate of linguistic diversity and inclusion in the NLP world. In *Proceedings of the 58th Annual Meeting of the Association for Computational Linguistics*, pages 6282–6293, Online. Association for Computational Linguistics.

Raviraj Joshi, Kanishk Singla, Anusha Kamath, Raunak Kalani, Rakesh Paul, Utkarsh Vaidya, Sanjay Singh Chauhan, Niranjan Wartikar, and Eileen Long. 2025. Adapting multilingual LLMs to low-resource languages using continued pre-training and synthetic corpus: A case study for Hindi LLMs. In *Proceedings of the First Workshop on Natural Language Processing for Indo-Aryan and Dravidian Languages*, pages 50–57, Abu Dhabi. Association for Computational Linguistics.

Jared Kaplan, Sam McCandlish, Tom Henighan, Tom B. Brown, Benjamin Chess, Rewon Child, Scott Gray, Alec Radford, Jeffrey Wu, and Dario Amodei. 2020. Scaling laws for neural language models.

Mohammed Safi Ur Rahman Khan, Priyam Mehta, Ananth Sankar, Umashankar Kumaravelan, Sumanth Doddapaneni, Suriyaprasaad B, Varun G, Sparsh Jain, Anoop Kunchukuttan, Pratyush Kumar, Raj Dabre, and Mitesh M. Khapra. 2024. IndicLLMSuite: A blueprint for creating pre-training and fine-tuning datasets for Indian languages. In *Proceedings of the 62nd Annual Meeting of the Association for Computational Linguistics (Volume 1: Long Papers)*, pages 15831–15879, Bangkok, Thailand. Association for Computational Linguistics.

Taku Kudo. 2018. Subword regularization: Improving neural network translation models with multiple subword candidates.

Viet Dac Lai, Nghia Ngo, Amir Pouran Ben Veyseh, Hieu Man, Franck Dernoncourt, Trung Bui, and Thien Huu Nguyen. 2023a. ChatGPT beyond English: Towards a comprehensive evaluation of large language models in multilingual learning. In *Findings of the Association for Computational Linguistics: EMNLP 2023*, pages 13171–13189, Singapore. Association for Computational Linguistics.

Viet Dac Lai, Chien Van Nguyen, Nghia Trung Ngo, Thuat Nguyen, Franck Dernoncourt, Ryan A. Rossi, and Thien Huu Nguyen. 2023b. Okapi: Instruction-tuned large language models in multiple languages with reinforcement learning from human feedback.

Anne Lauscher, Vinit Ravishankar, Ivan Vulić, and Goran Glavaš. 2020. From zero to hero: On the limitations of zero-shot language transfer with multilingual Transformers. In *Proceedings of the 2020 Conference on Empirical Methods in Natural Language Processing (EMNLP)*, pages 4483–4499, Online. Association for Computational Linguistics.

Xi Victoria Lin, Todor Mihaylov, Mikel Artetxe, Tianlu Wang, Shuohui Chen, Daniel Simig, Myle Ott, Naman Goyal, Shruti Bhosale, Jingfei Du, Ramakanth Pasunuru, Sam Shleifer, Punit Singh Koura, Vishrav Chaudhary, Brian O'Horo, Jeff Wang, Luke Zettlemoyer, Zornitsa Kozareva, Mona Diab, Veselin Stoyanov, and Xian Li. 2022. Few-shot learning with multilingual generative language models. In *Proceedings of the 2022 Conference on Empirical Methods in Natural Language Processing*, pages 9019–9052, Abu Dhabi, United Arab Emirates. Association for Computational Linguistics.

Ilya Loshchilov and Frank Hutter. 2019. Decoupled weight decay regularization. In *International Conference on Learning Representations*.

Yinquan Lu, Wenhao Zhu, Lei Li, Yu Qiao, and Fei Yuan. 2024. LLaMAX: Scaling linguistic horizons of LLM by enhancing translation capabilities beyond 100 languages. In *Findings of the Association for Computational Linguistics: EMNLP 2024*, pages 10748–10772, Miami, Florida, USA. Association for Computational Linguistics.

Yash Madhani, Mitesh M. Khapra, and Anoop Kunchukuttan. 2023. Bhasa-Abhijnaanam: Native-script and romanized language identification for 22 Indic languages. In *Proceedings of the 61st Annual Meeting of the Association for Computational Linguistics (Volume 2: Short Papers)*, pages 816–826, Toronto, Canada. Association for Computational Linguistics.

Kelly Marchisio, Wei-Yin Ko, Alexandre Berard, Théo Dehaze, and Sebastian Ruder. 2024. Understanding and mitigating language confusion in LLMs. In *Proceedings of the 2024 Conference on Empirical Methods in Natural Language Processing*, pages 6653–6677, Miami, Florida, USA. Association for Computational Linguistics.

Meta AI. 2024. Llama 3.2: Revolutionizing edge ai and vision with open, customizable models. https://ai.meta.com/blog/llama-3-2-connect-2024-vision-edge-mobile-devices/. Accessed: 2025-12-27.

Wilhelmina Nekoto, Vukosi Marivate, Tshinondiwa Matsila, Timi Fasubaa, Taiwo Fagbohungbe, Solomon Oluwole Akinola, Shamsuddeen Muhammad, Salomon Kabongo Kabenamualu, Salomey Osei, Freshia Sackey, Rubungo Andre Niyongabo, Ricky Macharm, Perez Ogayo, Orevaoghene Ahia, Musie Meressa Berhe, Mofetoluwa Adeyemi, Masabata Mokgesi-Selinga, Lawrence Okegbemi, Laura Martinus, Kolawole Tajudeen, Kevin Degila, Kelechi Ogueji, Kathleen Siminyu, Julia Kreutzer, Jason Webster, Jamiil Toure Ali, Jade Abbott, Iroro Orife, Ignatius Ezeani, Idris Abdulkadir Dangana, Herman Kamper, Hady Elsahar, Goodness Duru, Ghollah Kioko, Murhabazi Espoir, Elan van Biljon, Daniel Whitenack, Christopher Onyefuluchi, Chris Chinenye Emezue, Bonaventure F. P. Dossou, Blessing Sibanda, Blessing Bassey, Ayodele Olabiyi, Arshath Ramkilowan, Alp Öktem, Adewale Akinfaderin, and Abdallah Bashir. 2020. Participatory research for low-resourced machine translation: A case study in African languages. In *Findings of the Association for Computational Linguistics: EMNLP 2020*, pages 2144–2160, Online. Association for Computational Linguistics.

Hellina Hailu Nigatu, Atnafu Lambebo Tonja, Benjamin Rosman, Thamar Solorio, and Monojit Choudhury. 2024. The zeno's paradox of 'low-resource' languages. In *Proceedings of the 2024 Conference on Empirical Methods in Natural Language Processing*, pages 17753–17774, Miami, Florida, USA. Association for Computational Linguistics.

Pedro Javier Ortiz Suárez, Benoît Sagot, and Laurent Romary. 2019. Asynchronous pipelines for processing huge corpora on medium to low resource infrastructures. Proceedings of the Workshop on Challenges in the Management of Large Corpora (CMLC-7) 2019. Cardiff, 22nd July 2019, pages 9 – 16, Mannheim. Leibniz-Institut für Deutsche Sprache.

Jonas Pfeiffer, Ivan Vulić, Iryna Gurevych, and Sebastian Ruder. 2021. UNKs everywhere: Adapting multilingual language models to new scripts. In *Proceedings of the 2021 Conference on Empirical Methods in Natural Language Processing*, pages 10186–10203, Online and Punta Cana, Dominican Republic. Association for Computational Linguistics.

Jan Pfister, Julia Wunderle, and Andreas Hotho. 2025. LLäMmlein: Transparent, compact and competitive German-only language models from scratch. In *Proceedings of the 63rd Annual Meeting of the Association for Computational Linguistics (Volume 1: Long Papers)*, pages 2227–2246, Vienna, Austria. Association for Computational Linguistics.

Edoardo Maria Ponti, Goran Glavaš, Olga Majewska, Qianchu Liu, Ivan Vulić, and Anna Korhonen. 2020. XCOPA: A multilingual dataset for causal commonsense reasoning. In *Proceedings of the 2020 Conference on Empirical Methods in Natural Language Processing (EMNLP)*, pages 2362–2376, Online. Association for Computational Linguistics.

Ofir Press and Lior Wolf. 2017. Using the output embedding to improve language models. In *Proceedings of the 15th Conference of the European Chapter of the Association for Computational Linguistics: Volume 2, Short Papers*, pages 157–163, Valencia, Spain. Association for Computational Linguistics.

Alec Radford, Jeff Wu, Rewon Child, David Luan, Dario Amodei, and Ilya Sutskever. 2019. Language models are unsupervised multitask learners.

Phillip Rust, Jonas Pfeiffer, Ivan Vulić, Sebastian Ruder, and Iryna Gurevych. 2021. How good is your tokenizer? on the monolingual performance of multilingual language models. In *Proceedings of the 59th Annual Meeting of the Association for Computational Linguistics and the 11th International Joint Conference on Natural Language Processing (Volume 1: Long Papers)*, pages 3118–3135, Online. Association for Computational Linguistics.

Sarvam2B. 2024. sarvamai/sarvam-2b-v0.5 ů Hugging Face — huggingface.co. https://huggingface.co/sarvamai/sarvam-2b-v0.5. [Accessed 15-09-2024].

SarvamAI. 2023. Openhathi series. Accessed: 2025-01-12.

Lisa Schut, Yarin Gal, and Sebastian Farquhar. 2025. Do multilingual LLMs think in english? In *ICLR 2025 Workshop on Building Trust in Language Models and Applications*.

Lane Schwartz. 2022. Primum Non Nocere: Before working with Indigenous data, the ACL must confront ongoing colonialism. In *Proceedings of the 60th Annual Meeting of the Association for Computational Linguistics (Volume 2: Short Papers)*, pages 724–731, Dublin, Ireland. Association for Computational Linguistics.

Rico Sennrich, Barry Haddow, and Alexandra Birch. 2016. Neural machine translation of rare words with subword units. In *Proceedings of the 54th Annual Meeting of the Association for Computational Linguistics (Volume 1: Long Papers)*, pages 1715–1725, Berlin, Germany. Association for Computational Linguistics.

Sheikh Shafayat, Eunsu Kim, Juhyun Oh, and Alice Oh. 2024. Multi-fact: Assessing factuality of multilingual llms using factscore.

Noam Shazeer. 2020. Glu variants improve transformer.

Freda Shi, Mirac Suzgun, Markus Freitag, Xuezhi Wang, Suraj Srivats, Soroush Vosoughi, Hyung Won Chung, Yi Tay, Sebastian Ruder, Denny Zhou, Dipanjan Das, and Jason Wei. 2023. Language models are multilingual chain-of-thought reasoners. In *The Eleventh International Conference on Learning Representations*.

Oleh Shliazhko, Alena Fenogenova, Maria Tikhonova, Anastasia Kozlova, Vladislav Mikhailov, and Tatiana Shavrina. 2024. mGPT: Few-shot learners go multilingual. *Transactions of the Association for Computational Linguistics*, 12:58–79.

Anil Kumar Singh. 2008. Natural language processing for less privileged languages: Where do we come from? where are we going? In *International Joint Conference on Natural Language Processing*.

Jianlin Su, Yu Lu, Shengfeng Pan, Ahmed Murtadha, Bo Wen, and Yunfeng Liu. 2022. Roformer: Enhanced transformer with rotary position embedding.

Lintang Sutawika, Hailey Schoelkopf, Leo Gao, Stella Biderman, Baber Abbasi, Jonathan Tow, ben fattori, Charles Lovering, farzanehnakhaee70, Jason Phang, Anish Thite, Fazz, Aflah, Niklas Muennighoff, Thomas Wang, sdtblck, gakada, nopperl, researcher2, tttyuntian, Chris, Julen Etxaniz, Zdenk Kasner, Khalid, Jeffrey Hsu, Hanwool Albert Lee, Anjor Kanekar, AndyZwei, Pawan Sasanka Ammanamanchi, and Dirk Groeneveld. 2024. Eleutherai/lm-evaluation-harness: v0.4.1.

Rohan Taori, Ishaan Gulrajani, Tianyi Zhang, Yann Dubois, Xuechen Li, Carlos Guestrin, Percy Liang, and Tatsunori B. Hashimoto. 2023. Stanford alpaca: An instruction-following llama model. https://github.com/tatsu-lab/stanford_alpaca.

Gemma Team, Morgane Riviere, Shreya Pathak, Pier Giuseppe Sessa, Cassidy Hardin, Surya Bhupatiraju, Léonard Hussenot, Thomas Mesnard, Bobak Shahriari, Alexandre Ramé, Johan Ferret, Peter Liu, Pouya Tafti, Abe Friesen, Michelle Casbon, Sabela Ramos, Ravin Kumar, Charline Le Lan, Sammy Jerome, Anton Tsitsulin, Nino Vieillard, Piotr Stanczyk, Sertan Girgin, Nikola Momchev, Matt Hoffman, Shantanu Thakoor, Jean-Bastien Grill, Behnam Neyshabur, Olivier Bachem, Alanna Walton, Aliaksei Severyn, Alicia Parrish, Aliya Ahmad, Allen Hutchison, Alvin Abdagic, Amanda Carl, Amy Shen, Andy Brock, Andy Coenen, Anthony Laforge, Antonia Paterson, Ben Bastian, Bilal Piot, Bo Wu, Brandon Royal, Charlie Chen, Chintu Kumar, Chris Perry, Chris Welty, Christopher A. Choquette-Choo, Danila Sinopalnikov, David Weinberger, Dimple Vijaykumar, Dominika Rogoziska, Dustin Herbison, Elisa Bandy, Emma Wang, Eric Noland, Erica Moreira, Evan Senter, Evgenii Eltyshev, Francesco Visin, Gabriel Rasskin, Gary Wei, Glenn Cameron, Gus Martins, Hadi Hashemi, Hanna Klimczak-Pluciska, Harleen Batra, Harsh Dhand, Ivan Nardini, Jacinda Mein, Jack Zhou, James Svensson, Jeff Stanway, Jetha Chan, Jin Peng Zhou, Joana Carrasqueira, Joana Iljazi, Jocelyn Becker, Joe Fernandez, Joost van Amersfoort, Josh Gordon, Josh Lipschultz, Josh Newlan, Ju yeong Ji, Kareem Mohamed, Kartikeya Badola, Kat Black, Katie Millican, Keelin McDonell, Kelvin Nguyen, Kiranbir Sodhia, Kish Greene, Lars Lowe Sjoesund, Lauren Usui, Laurent Sifre, Lena Heuermann, Leticia Lago, Lilly McNealus, Livio Baldini Soares, Logan Kilpatrick, Lucas Dixon, Luciano Martins, Machel Reid, Manvinder Singh, Mark Iverson, Martin Görner, Mat Velloso, Mateo Wirth, Matt Davidow, Matt Miller, Matthew Rahtz, Matthew Watson, Meg Risdal, Mehran Kazemi, Michael Moynihan, Ming Zhang, Minsuk Kahng, Minwoo Park, Mofi Rahman, Mohit Khatwani, Natalie Dao, Nenshad Bardoliwalla, Nesh Devanathan, Neta Dumai, Nilay Chauhan, Oscar Wahltinez, Pankil Botarda, Parker Barnes, Paul Barham, Paul Michel, Pengchong Jin, Petko Georgiev, Phil Culliton, Pradeep Kuppala, Ramona Comanescu, Ramona Merhej, Reena Jana, Reza Ardeshir Rokni, Rishabh Agarwal, Ryan Mullins, Samaneh Saadat, Sara Mc Carthy, Sarah Perrin, Sébastien M. R. Arnold, Sebastian Krause, Shengyang Dai, Shruti Garg, Shruti Sheth, Sue Ronstrom, Susan Chan, Timothy Jordan, Ting Yu, Tom Eccles, Tom Hennigan, Tomas Kocisky, Tulsee Doshi, Vihan Jain, Vikas Yadav, Vilobh Meshram, Vishal Dharmadhikari, Warren Barkley, Wei Wei, Wenming Ye, Woohyun Han, Woosuk Kwon, Xiang Xu, Zhe Shen, Zhitao Gong, Zichuan Wei, Victor Cotruta, Phoebe Kirk, Anand Rao, Minh Giang, Ludovic Peran, Tris Warkentin, Eli Collins, Joelle Barral, Zoubin Ghahramani, Raia Hadsell, D. Sculley, Jeanine Banks, Anca Dragan, Slav Petrov, Oriol Vinyals, Jeff Dean, Demis Hassabis, Koray Kavukcuoglu, Clement Farabet, Elena Buchatskaya, Sebastian Borgeaud, Noah Fiedel, Armand Joulin, Kathleen Kenealy, Robert Dadashi, and Alek Andreev. 2024. Gemma 2: Improving open language models at a practical size.

NLLB Team, Marta R. Costa-jussà, James Cross, Onur Çelebi, Maha Elbayad, Kenneth Heafield, Kevin Heffernan, Elahe Kalbassi, Janice Lam, Daniel Licht, Jean Maillard, Anna Sun, Skyler Wang, Guillaume Wenzek, Al Youngblood, Bapi Akula, Loic Barrault, Gabriel Mejia Gonzalez, Prangthip Hansanti, John Hoffman, Semarley Jarrett, Kaushik Ram Sadagopan, Dirk Rowe, Shannon Spruit, Chau Tran, Pierre Andrews, Necip Fazil Ayan, Shruti Bhosale, Sergey Edunov, Angela Fan, Cynthia Gao, Vedanuj Goswami, Francisco Guzmán, Philipp Koehn, Alexandre Mourachko, Christophe Ropers, Safiyyah Saleem, Holger Schwenk, and Jeff Wang. 2022. No language left behind: Scaling human-centered machine translation.

Telugu-LLM-Labs. 2025. Indic-gemma-2b-finetuned-sft-navarasa-2.0. Accessed: 2025-03-09.

Hugo Touvron, Thibaut Lavril, Gautier Izacard, Xavier Martinet, Marie-Anne Lachaux, Timothée Lacroix, Baptiste Rozière, Naman Goyal, Eric Hambro, Faisal Azhar, Aurelien Rodriguez, Armand Joulin, Edouard Grave, and Guillaume Lample. 2023. Llama: Open and efficient foundation language models.

Yulia Tsvetkov. 2017. Opportunities and challenges in working with low-resource languages. *Slides Part-1*, 2.

Karthik Valmeekam, Matthew Marquez, Alberto Olmo, Sarath Sreedharan, and Subbarao Kambhampati. 2024. Planbench: an extensible benchmark for evaluating large language models on planning and reasoning about change. In *Proceedings of the 37th International Conference on Neural Information Processing Systems*, NIPS '23, Red Hook, NY, USA. Curran Associates Inc.

Ashish Vaswani, Noam Shazeer, Niki Parmar, Jakob Uszkoreit, Llion Jones, Aidan N. Gomez, Łukasz Kaiser, and Illia Polosukhin. 2017. Attention is all you need. In *Proceedings of the 31st International Conference on Neural Information Processing Systems*, NIPS'17, page 60006010, Red Hook, NY, USA. Curran Associates Inc.

Yizhong Wang, Yeganeh Kordi, Swaroop Mishra, Alisa Liu, Noah A. Smith, Daniel Khashabi, and Hannaneh Hajishirzi. 2023. Self-instruct: Aligning language models with self-generated instructions.

BigScience Workshop, :, Teven Le Scao, Angela Fan, Christopher Akiki, Ellie Pavlick, Suzana Ili, Daniel Hesslow, Roman Castagné, Alexandra Sasha Luccioni, François Yvon, Matthias Gallé, Jonathan

Tow, Alexander M. Rush, Stella Biderman, Albert Webson, Pawan Sasanka Ammanamanchi, Thomas Wang, Benoît Sagot, Niklas Muennighoff, Albert Villanova del Moral, Olatunji Ruwase, Rachel Bawden, Stas Bekman, Angelina McMillan-Major, Iz Beltagy, Huu Nguyen, Lucile Saulnier, Samson Tan, Pedro Ortiz Suarez, Victor Sanh, Hugo Laurençon, Yacine Jernite, Julien Launay, Margaret Mitchell, Colin Raffel, Aaron Gokaslan, Adi Simhi, Aitor Soroa, Alham Fikri Aji, Amit Alfassy, Anna Rogers, Ariel Kreisberg Nitzav, Canwen Xu, Chenghao Mou, Chris Emezue, Christopher Klamm, Colin Leong, Daniel van Strien, David Ifeoluwa Adelani, Dragomir Radev, Eduardo González Ponferrada, Efrat Levkovizh, Ethan Kim, Eyal Bar Natan, Francesco De Toni, Gérard Dupont, Germán Kruszewski, Giada Pistilli, Hady Elsahar, Hamza Benyamina, Hieu Tran, Ian Yu, Idris Abdulmumin, Isaac Johnson, Itziar Gonzalez-Dios, Javier de la Rosa, Jenny Chim, Jesse Dodge, Jian Zhu, Jonathan Chang, Jörg Frohberg, Joseph Tobing, Joydeep Bhattacharjee, Khalid Almubarak, Kimbo Chen, Kyle Lo, Leandro Von Werra, Leon Weber, Long Phan, Loubna Ben allal, Ludovic Tanguy, Manan Dey, Manuel Romero Muñoz, Maraim Masoud, María Grandury, Mario ako, Max Huang, Maximin Coavoux, Mayank Singh, Mike Tian-Jian Jiang, Minh Chien Vu, Mohammad A. Jauhar, Mustafa Ghaleb, Nishant Subramani, Nora Kassner, Nurulaqilla Khamis, Olivier Nguyen, Omar Espejel, Ona de Gibert, Paulo Villegas, Peter Henderson, Pierre Colombo, Priscilla Amuok, Quentin Lhoest, Rheza Harliman, Rishi Bommasani, Roberto Luis López, Rui Ribeiro, Salomey Osei, Sampo Pyysalo, Sebastian Nagel, Shamik Bose, Shamsuddeen Hassan Muhammad, Shanya Sharma, Shayne Longpre, Somaieh Nikpoor, Stanislav Silberberg, Suhas Pai, Sydney Zink, Tiago Timponi Torrent, Timo Schick, Tristan Thrush, Valentin Danchev, Vassilina Nikoulina, Veronika Laippala, Violette Lepercq, Vrinda Prabhu, Zaid Alyafeai, Zeerak Talat, Arun Raja, Benjamin Heinzerling, Chenglei Si, Davut Emre Taar, Elizabeth Salesky, Sabrina J. Mielke, Wilson Y. Lee, Abheesht Sharma, Andrea Santilli, Antoine Chaffin, Arnaud Stiegler, Debajyoti Datta, Eliza Szczechla, Gunjan Chhablani, Han Wang, Harshit Pandey, Hendrik Strobelt, Jason Alan Fries, Jos Rozen, Leo Gao, Lintang Sutawika, M Saiful Bari, Maged S. Al-shaibani, Matteo Manica, Nihal Nayak, Ryan Teehan, Samuel Albanie, Sheng Shen, Srulik Ben-David, Stephen H. Bach, Taewoon Kim, Tali Bers, Thibault Fevry, Trishala Neeraj, Urmish Thakker, Vikas Raunak, Xiangru Tang, Zheng-Xin Yong, Zhiqing Sun, Shaked Brody, Yallow Uri, Hadar Tojarieh, Adam Roberts, Hyung Won Chung, Jaesung Tae, Jason Phang, Ofir Press, Conglong Li, Deepak Narayanan, Hatim Bourfoune, Jared Casper, Jeff Rasley, Max Ryabinin, Mayank Mishra, Minjia Zhang, Mohammad Shoeybi, Myriam Peyrounette, Nicolas Patry, Nouamane Tazi, Omar Sanseviero, Patrick von Platen, Pierre Cornette, Pierre François Lavallée, Rémi Lacroix, Samyam Rajbhandari, Sanchit Gandhi, Shaden Smith, Stéphane Requena, Suraj Patil, Tim Dettmers, Ahmed Baruwa, Amanpreet Singh, Anastasia Cheveleva, Anne-Laure Ligozat, Arjun Subramonian, Aurélie Névéol, Charles Lovering, Dan Garrette, Deepak Tunuguntla, Ehud Reiter, Ekaterina Taktasheva, Ekaterina Voloshina, Eli Bogdanov, Genta Indra Winata, Hailey Schoelkopf, Jan-Christoph Kalo, Jekaterina Novikova, Jessica Zosa Forde, Jordan Clive, Jungo Kasai, Ken Kawamura, Liam Hazan, Marine Carpuat, Miruna Clinciu, Najoung Kim, Newton Cheng, Oleg Serikov, Omer Antverg, Oskar van der Wal, Rui Zhang, Ruochen Zhang, Sebastian Gehrmann, Shachar Mirkin, Shani Pais, Tatiana Shavrina, Thomas Scialom, Tian Yun, Tomasz Limisiewicz, Verena Rieser, Vitaly Protasov, Vladislav Mikhailov, Yada Pruksachatkun, Yonatan Belinkov, Zachary Bamberger, Zdenk Kasner, Alice Rueda, Amanda Pestana, Amir Feizpour, Ammar Khan, Amy Faranak, Ana Santos, Anthony Hevia, Antigona Unldreaj, Arash Aghagol, Arezoo Abdollahi, Aycha Tammour, Azadeh HajiHosseini, Bahareh Behroozi, Benjamin Ajibade, Bharat Saxena, Carlos Muñoz Ferrandis, Daniel McDuff, Danish Contractor, David Lansky, Davis David, Douwe Kiela, Duong A. Nguyen, Edward Tan, Emi Baylor, Ezinwanne Ozoani, Fatima Mirza, Frankline Ononiwu, Habib Rezanejad, Hessie Jones, Indrani Bhattacharya, Irene Solaiman, Irina Sedenko, Isar Nejadgholi, Jesse Passmore, Josh Seltzer, Julio Bonis Sanz, Livia Dutra, Mairon Samagaio, Maraim Elbadri, Margot Mieskes, Marissa Gerchick, Martha Akinlolu, Michael McKenna, Mike Qiu, Muhammed Ghauri, Mykola Burynok, Nafis Abrar, Nazneen Rajani, Nour Elkott, Nour Fahmy, Olanrewaju Samuel, Ran An, Rasmus Kromann, Ryan Hao, Samira Alizadeh, Sarmad Shubber, Silas Wang, Sourav Roy, Sylvain Viguier, Thanh Le, Tobi Oyebade, Trieu Le, Yoyo Yang, Zach Nguyen, Abhinav Ramesh Kashyap, Alfredo Palasciano, Alison Callahan, Anima Shukla, Antonio Miranda-Escalada, Ayush Singh, Benjamin Beilharz, Bo Wang, Caio Brito, Chenxi Zhou, Chirag Jain, Chuxin Xu, Clémentine Fourrier, Daniel León Periñán, Daniel Molano, Dian Yu, Enrique Manjavacas, Fabio Barth, Florian Fuhrimann, Gabriel Altay, Giyaseddin Bayrak, Gully Burns, Helena U. Vrabec, Imane Bello, Ishani Dash, Jihyun Kang, John Giorgi, Jonas Golde, Jose David Posada, Karthik Rangasai Sivaraman, Lokesh Bulchandani, Lu Liu, Luisa Shinzato, Madeleine Hahn de Bykhovetz, Maiko Takeuchi, Marc Pàmies, Maria A Castillo, Marianna Nezhurina, Mario Sänger, Matthias Samwald, Michael Cullan, Michael Weinberg, Michiel De Wolf, Mina Mihaljcic, Minna Liu, Moritz Freidank, Myungsun Kang, Natasha Seelam, Nathan Dahlberg, Nicholas Michio Broad, Nikolaus Muellner, Pascale Fung, Patrick Haller, Ramya Chandrasekhar, Renata Eisenberg, Robert Martin, Rodrigo Canalli, Rosaline Su, Ruisi Su, Samuel Cahyawijaya, Samuele Garda, Shlok S Deshmukh, Shubhanshu Mishra, Sid Kiblawi, Simon Ott, Sinee Sang-aroonsiri, Srishti Kumar, Stefan Schweter, Sushil Bharati, Tanmay Laud, Théo Gigant, Tomoya Kainuma, Woj-ciech Kusa, Yanis Labrak, Yash Shailesh Bajaj, Yash Venkatraman, Yifan Xu, Yingxin Xu, Yu Xu, Zhe Tan, Zhongli Xie, Zifan Ye, Mathilde Bras, Younes Belkada, and Thomas Wolf. 2023. Bloom: A 176b-parameter open-access multilingual language model.

Linting Xue, Noah Constant, Adam Roberts, Mihir Kale, Rami Al-Rfou, Aditya Siddhant, Aditya Barua, and Colin Raffel. 2021. mT5: A massively multilingual pre-trained text-to-text transformer. In *Proceedings of the 2021 Conference of the North American Chapter of the Association for Computational Linguistics: Human Language Technologies*, pages 483–498, Online. Association for Computational Linguistics.

Greg Yang, Edward J Hu, Igor Babuschkin, Szymon Sidor, Xiaodong Liu, David Farhi, Nick Ryder, Jakub Pachocki, Weizhu Chen, and Jianfeng Gao. 2021. Tuning large neural networks via zero-shot hyperparameter transfer. In *Advances in Neural Information Processing Systems*.

Biao Zhang and Rico Sennrich. 2019. *Root mean square layer normalization*. Curran Associates Inc., Red Hook, NY, USA.

Susan Zhang, Stephen Roller, Naman Goyal, Mikel Artetxe, Moya Chen, Shuohui Chen, Christopher Dewan, Mona Diab, Xian Li, Xi Victoria Lin, Todor Mihaylov, Myle Ott, Sam Shleifer, Kurt Shuster, Daniel Simig, Punit Singh Koura, Anjali Sridhar, Tianlu Wang, and Luke Zettlemoyer. 2022. Opt: Open pre-trained transformer language models.

Wenzhen Zheng, Wenbo Pan, Xu Xu, Libo Qin, Li Yue, and Ming Zhou. 2024a. Breaking language barriers: Cross-lingual continual pre-training at scale. In *Proceedings of the 2024 Conference on Empirical Methods in Natural Language Processing*, pages 7725–7738, Miami, Florida, USA. Association for Computational Linguistics.

Yaowei Zheng, Richong Zhang, Junhao Zhang, Yanhan Ye, and Zheyan Luo. 2024b. LlamaFactory: Unified efficient fine-tuning of 100+ language models. In *Proceedings of the 62nd Annual Meeting of the Association for Computational Linguistics (Volume 3: System Demonstrations)*, pages 400–410, Bangkok, Thailand. Association for Computational Linguistics.

# Appendix

## A Tokenizers

Table 7 illustrates the tokenization behavior of our hybrid tokenizer compared to standard BPE across several Indian languages. Figure 3 compares the Llama-2 tokenizer vs our multilingual tokenizer for Indian languages.

## B Annotators Information and Guidelines to design Bengali Instruction-Tuning Dataset

The human-authored Bangla 5K instruction dataset was created by 20 native Bangla-speaking annotators, each responsible for generating 250 instruction-response pairs, resulting in 5,000 high-quality samples. All annotators have prior experience in linguistic annotation, translation, or writing. Each instruction-response pair underwent peer review for linguistics and factual correctness, fluency, clarity, and cultural appropriateness. Agreement was measured using Fleiss Kappa across all annotators, yielding an average score of 0.87, indicating strong reliability and consistency in annotation. Due to budget constraints, manual post-editing was performed only for Hindi and Marathi. Word error rates averaged 8% for Hindi and 13% for Marathi, and these errors were manually corrected by annotators following grammatical correction guidelines.

We construct a manually curated dataset of 5,000 high-quality instructionresponse pairs in Bengali, focusing on cultural, literary, and practical domains relevant to Bengali-speaking users in India. Below, we outline the core annotation guidelines used to ensure consistency, linguistic accuracy, and cultural relevance.

**Annotation Protocol**

Each data point consists of:

- **Instruction:** A natural prompt in Bengali, emulating user queries or tasks.
- **Response:** A helpful, accurate, and contextually appropriate answer in Bengali.
- **Category (optional):** Domain label (e.g., Literature, Daily Life).

**Domain Coverage.**

To promote diversity, we target a balanced distribution across ten culturally salient domains: literature, culture/tradition, history/politics, religion, education, health and daily life, technology, ethics, creative writing, and casual conversation.

| Text | BPE | BPE+Unigram |
|---|---|---|
| কিংবদন্তি (kiṃbadanti) | ['_কিং', 'ব', 'দ', 'ন্ত', 'ি'] | ['_কিং', 'ব', 'দ', 'ন্তি'] |
| নিয়মানুবর্তিতা (niẏamānubartitā) | :['_নিয়', 'মান', 'ু', 'বর্ত', 'িতা'] | ['_নিয়', 'মা', 'নু', 'বর্ত', 'িতা'] |
| পরিবারের (paribārera) | ['_পরি', 'বারের'] | ['_পরিবার', 'ের'] |
| স্বাচ্ছন্দ্য (sbācchandya) | :['_স্বা', 'চ্ছ', 'ন্দ', '্য'] | :['_স্বা', 'চ্ছ', 'ন্দ', '্য'] |
| ব্যক্তিগত (byaktigata) | :['_ব্যক্তি', 'গত'] | :['_ব্যক্তি', 'গত'] |
| பயணித்தார்கள் (payaṇittārkaḷ) | ['_பயணி', 'த்த', 'ார்கள்'] | ['_பயணித்த', 'ார்கள்'] |
| எழுதிக்கொண்டேபோயிருந்தார்கள் (ĕḻutikkŏṇṭepoyiruntārkaḷ) | ['_எழு', 'திக்க', 'ொண்ட', 'ே', 'போ', 'ய', 'ிருந்த', 'ார்கள்'] | ['_எழு', 'திக்க', 'ொண்ட', 'ே', 'போ', 'யிருந்த', 'ார்கள்'] |
| చదువుకుంటూనేవున్నవారికోసం (caduvukuṃṭūnevunnavārikosaṃ) | ['_చదువు', 'కు', 'ంటూ', 'నే', 'వు', 'న్న', 'వార', 'ిక', 'ో', 'సం'] | ['_చదువు', 'కు', 'ంటూ', 'నే', 'వ', 'ున్న', 'వార', 'ిక', 'ో', 'సం'] |
| लिहूनघेतलेल्यानुसारही (lihūnaghetalelyānusārahī) | ['_लि', 'हून', 'घ', 'ेत', 'लेल्या', 'नुसार', 'ही'] | ['_लिह', 'ून', 'घे', 'तले', 'ल्या', 'नुसार', 'ही'] |
| लिखकरदेनेवालोंकेलिएभी (likhakaradenevāloṃkeliebhī) | ['_लिखकर', 'दे', 'ने', 'वालों', 'के', 'लिए', 'भी'] | ['_लिख', 'कर', 'दे', 'ने', 'वाले', 'ओं', 'के', 'लिए', 'भी'] |

Table 7: Comparison of BPE and BPE+Unigram Tokenization for 5 Indian languages.

Test input indic multilingual text in 7 languages: ভাৰত মোৰ মাতৃভূমি ভারত আমার মাতৃভূমি साडा देश भारत है भारत हो म्हजो देश भारत हमर देश अछि ଭାରତ ହେଉଛି ମୋର ମାତୃଭୂମି अस्माकं देशः भारतवर्षः

Tokenized by LLaMA-2 32k tokenizer:['_', '<0xE0>', '<0xA6>', '<0xAD>', 'া', '<0xE0>', '<0xA7>', '<0xB0>', 'ত', '_', 'ম', 'ো', '<0xE0>', '<0xA7>', '<0xB0>', '_', 'ম', 'া', 'ত', '<0xE0>', '<0xA7>', '<0x83>', '<0xE0>', '<0xA6>', '<0xAD>', '<0xE0>', '<0xA7>', '<0x82>', 'ম', 'ি', '_', '<0xE0>', '<0xA6>', '<0xAD>', 'া', 'র', 'ত', '_', '<0xE0>', '<0xA6>', '<0x86>', 'ম', 'া', 'র', '_', 'ম', 'া', 'ত', '<0xE0>', '<0xA7>', '<0x83>', '<0xE0>', '<0xA6>', '<0xAD>', '<0xE0>', '<0xA7>', '<0x82>', 'ম', 'ি', '_', 'स', 'ा', 'ड', 'ा', '_', 'द', 'े', 'श', '_', 'भ', 'ा', 'र', 'त', '_', 'ह', 'ै', '_', 'भ', 'ा', 'र', 'त', '_', 'ह', 'ो', '_', 'म', '्', 'ह', 'ज', 'ो', '_', 'द', 'े', 'श', '_', 'भ', 'ा', 'र', 'त', '_', 'ह', 'म', 'र', '_', 'द', 'े', 'श', '_', 'अ', '<0xE0>', '<0xA4>', '<0x9B>', 'ि', '_', '<0xE0>', '<0xAC>', '<0xAD>', '<0xE0>', '<0xAC>', '<0xBE>', '<0xE0>', '<0xAC>', '<0xB0>', '<0xE0>', '<0xAC>', '<0xA4>', '_', '<0xE0>', '<0xAC>', '<0xB9>', '<0xE0>', '<0xAD>', '<0x87>', '<0xE0>', '<0xAC>', '<0x89>', '<0xE0>', '<0xAC>', '<0x9B>', '<0xE0>', '<0xAC>', '<0xBF>', '_', '<0xE0>', '<0xAC>', '<0xAE>', '<0xE0>', '<0xAD>', '<0x8B>', '<0xE0>', '<0xAC>', '<0xB0>', '_', '<0xE0>', '<0xAC>', '<0xAE>', '<0xE0>', '<0xAC>', '<0xBE>', '<0xE0>', '<0xAC>', '<0xA4>', '<0xE0>', '<0xAD>', '<0x83>', '<0xE0>', '<0xAC>', '<0xAD>', '<0xE0>', '<0xAD>', '<0x82>', '<0xE0>', '<0xAC>', '<0xAE>', '<0xE0>', '<0xAC>', '<0xBF>', '_', 'अ', 'स', '्', 'म', 'ा', 'क', 'ं', '_', 'द', 'े', 'श', '<0xE0>', '<0xA4>', '<0x83>', '_', 'भ', 'ा', 'र', 'त', 'व', 'र', '्', 'ष', '<0xE0>', '<0xA4>', '<0x83>']

Tokenized by mBharat tokenizer:['_ভাৰত', '_মোৰ', '_মাতৃ', 'ভূ', 'মি', '_ভারত', '_আমার', '_মাতৃ', 'ভূ', 'মি', '_सा', 'डा', '_देश', '_भारत', '_है', '_भारत', '_हो', '_म्ह', 'जो', '_देश', '_भारत', '_हम', 'र', '_देश', '_अछि', '_ଭାରତ', '_ହେଉଛି', '_ମୋର', '_ମ', 'ାତ', 'ୃ', 'ଭୂମି', '_अ', 'स्मा', 'कं', '_देश', 'ः', '_भारत', 'वर्ष', 'ः']

Figure 3: mBharat Tokenizer v/s LLaMa-2 tokenizer for input in 7 Indian languages.

| Language | Avg. Instruction Length | Avg. Input. Length | Avg. Response Length |
|---|---|---|---|
| Bangla | $63.7 \pm 61.47$ | $95.36 \pm 468.49$ | $505.45 \pm 499.93$ |
| Hindi | $63.54 \pm 63.28$ | $97.31 \pm 478.18$ | $461.89 \pm 462.19$ |
| Marathi | $58.91 \pm 65.08$ | $95.73 \pm 470.83$ | $499.30 \pm 478.70$ |
| Tamil | $72.72 \pm 66.83$ | $109.33 \pm 534.16$ | $496.96 \pm 456.53$ |
| Telugu | $64.69 \pm 63.07$ | $99.82 \pm 488.04$ | $458.20 \pm 419.08$ |

Table 8: Average # tokens in each Instruction, Input, and Response across 5 Indian languages for the instruction-tuning datasets.

**Instruction Types.**

We include a variety of prompt formats, including:

- Informational Q&A (e.g., historical facts, literary interpretation)
- How-to and procedural instructions
- Creative writing prompts (e.g., poem, story)
- Comparative and opinion-based prompts
- Grammar correction and translation tasks

**Language Guidelines.**

Instructions and responses are written in standard colloquial Bengali. Annotators avoid code-switching and maintain gender neutrality unless contextually necessary. All content is free of offensive, biased, or politically sensitive material.

**Quality Control**

All samples undergo dual-stage validation:

1. Peer review for factual correctness, fluency, and clarity.
2. Final check for linguistic naturalness and domain balance.

This dataset aims to support instruction-tuning of LLMs in Bengali by providing diverse, culturally grounded, and linguistically accurate examples.

## C Added Evaluation

### C.1 Language Specific Evaluation

In this subsection, we discuss language-specific benchmarks for five Indian languages: Bengali, Hindi, Marathi, Tamil, and Telugu. Tables 9, 10, 11, 12, 13 shows the respective language specific quantitative benchmark results.

| Models | Size | MMLU-Bangla | ARC-Bangla | Belebele-Bangla | Average (Bangla) | Belebele-Assamese |
|---|---|---|---|---|---|---|
| Paramanu-Bangla (ours) | 108M | 24.82 | 25.75 | 25.11 | 25.22 | 25.33 |
| Paramanu-Bangla-instruct (ours) | 108M | **27.60** | **28.50** | **32.45** | **29.52** | **30.54** |
| mParamanu (ours) | 92M* | 25.78 | 26.18 | 22.44 | 24.80 | 22.44 |
| mParamanu (ours) | 162M | 25.29 | 26.01 | 27.44 | 26.24 | 29.00 |
| mParamanu (ours) | 237M* | 22.52 | 25.92 | 24.44 | 24.29 | 24.28 |
| BanglaT5 | 247M | 22.62 | 20.27 | 22.89 | 21.92 | 22.89 |
| BanglaByT5 | 300M | 24.23 | 17.37 | 22.00 | 21.20 | 21.78 |
| Bloom | 560M | 22.61 | 26.00 | 22.89 | 23.83 | 22.78 |
| Bloomz (instruction-tuned) | 560M | 25.82 | 23.43 | 22.77 | 24.01 | 25.11 |
| xGLM | 564M | 22.61 | 18.47 | 23.55 | 21.54 | 22.77 |
| Llama-3.2 | 1B | 25.83 | 20.44 | 28.44 | 24.90 | 28.33 |
| Bloom | 1.1B | 23.90 | 24.37 | 26.00 | 24.75 | 26.89 |
| Bloomz | 1.1B | 25.19 | 20.61 | 24.55 | 23.45 | 22.33 |
| mGPT | 1.3B | 24.12 | 18.90 | 25.55 | 22.86 | 24.88 |
| xGLM | 1.7B | 24.46 | 19.50 | 22.77 | 22.24 | 22.55 |
| Sarvam | 2B | 24.05 | 28.40 | 23.22 | 25.22 | 27.78 |
| Indic-Gemma-Navrasa (instruction-tuned) | 2B | **29.07** | 20.87 | 25.88 | 25.27 | 24.44 |
| xGLM | 2.9B | 23.75 | 19.85 | 23.22 | 22.27 | 22.66 |
| Llama-3.2 | 3B | **32.51** | 21.47 | **36.55** | **30.17** | **33.88** |
| xGLM | 4.5B | 25.74 | 21.04 | 21.88 | 22.88 | 22.44 |
| Bloom | 7B | 27.10 | 26.09 | 23.22 | 25.47 | 23.11 |
| Bloomz (instruction-tuned) | 7B | **32.46** | 27.20 | **53.67** | **37.77** | **48.00** |
| xGLM | 7.5B | 24.69 | 19.41 | 24.88 | 22.99 | 24.44 |
| Llama-3 | 8B | **35.77** | 23.86 | **41.00** | **33.54** | **35.77** |
| Aya23 | 8B | 26.94 | 18.81 | 31.55 | 25.76 | **31.22** |

* Trained for the same number of steps.

Table 9: Zero-shot evaluation of LLMs across translated benchmarks of MMLU, HellaSwag, ARC datasets, and Belebele in Bengali script. All benchmarks report Accuracy. Models that performed better than our models have been **underlined and bold**, the best performance of our model has been **bold**.

| Models | Size | MMLU-Hindi | HellaSwag-Hindi | ARC-Hindi | XStoryCloze-Hindi | XNLI-Hindi | Belebele-Hindi | Average |
|---|---|---|---|---|---|---|---|---|
| Paramanu-Sanskrit (ours) | 139M | 25.16 | 25.64 | 25.17 | 50.23 | 34.46 | 25.66 | 31.05 |
| mParamanu (ours) | 92M* | 23.94 | 25.31 | 26.28 | 47.05 | 33.45 | 24.44 | 30.07 |
| mParamanu (ours) | 162M* | 24.84 | 24.87 | 22.35 | 49.24 | 33.70 | 25.44 | 30.07 |
| mParamanu (ours) | 237M* | 22.78 | 25.17 | 27.57 | 46.79 | 33.13 | 21.89 | 29.55 |
| Paramanu-Marathi (ours) | 208M | 25.49 | 26.59 | 23.97 | 48.71 | 33.73 | 27.33 | 30.97 |
| Paramanu-Marathi-instruct (ours) | 208M | 23.71 | 27.78 | 23.89 | 50.89 | 34.10 | 22.89 | 30.54 |
| Paramanu-Hindi (ours) | 162M* | 23.15 | 25.37 | 27.31 | 48.91 | 33.17 | 22.67 | 30.09 |
| Paramanu-Hindi (ours) | 367M* | 24.38 | 24.83 | 27.05 | 47.92 | 32.00 | 23.33 | 29.92 |
| Paramanu-Hindi (ours) | 367M | 25.18 | 25.02 | 27.14 | 48.78 | 33.49 | 26.22 | 30.97 |
| Paramanu-Hindi-instruct (ours) | 367M | **30.25** | **29.42** | **30.23** | **58.00** | **40.25** | **42.78** | **40.14** |
| Bloom | 560M | 23.67 | 27.50 | 23.88 | 54.79 | 40.80 | 26.44 | 32.84 |
| Bloomz (instruction-tuned) | 560M | 25.87 | 26.48 | 24.40 | 55.53 | 35.58 | 26.00 | 32.31 |
| xGLM | 564M | 22.70 | 26.96 | 20.20 | 52.00 | 38.31 | 24.00 | 30.69 |
| Llama-3.2 | 1B | 28.22 | 28.95 | 23.97 | 56.25 | 40.08 | 29.77 | 34.54 |
| Bloom | 1.1B | 23.86 | 28.28 | 24.74 | 55.59 | **42.77** | 28.00 | 33.87 |
| Bloomz | 1.1B | 24.90 | 28.54 | 21.06 | 56.65 | 37.87 | 22.22 | 31.87 |
| mGPT | 1.3B | 24.26 | 27.42 | 19.86 | 52.74 | **41.32** | 26.11 | 31.95 |
| xGLM | 1.7B | 24.70 | 28.46 | 20.63 | 55.79 | 38.99 | 24.00 | 32.09 |
| Sarvam | 2B | 24.54 | **33.66** | 28.00 | **60.29** | **46.74** | 24.44 | 36.27 |
| Indic-Gemma-Navrasa (instruction-tuned) | 2B | 29.63 | **30.31** | 22.43 | **60.62** | 37.22 | 26.22 | 34.40 |
| xGLM | 2.9B | 24.47 | 29.19 | 21.23 | 57.57 | **42.65** | 24.00 | 33.18 |
| Llama-3.2 | 3B | **34.88** | **32.78** | 24.65 | **60.75** | **43.45** | **43.88** | 40.06 |
| Nemotron-Hindi | 4B | **41.64** | **37.86** | **31.93** | **65.91** | **40.92** | **53.77** | **45.33** |
| xGLM | 4.5B | 25.93 | 28.44 | 21.06 | 56.84 | **41.28** | 23.44 | 32.83 |
| Bloom | 7B | 27.04 | **31.39** | 26.36 | **60.55** | **47.18** | 23.00 | 35.92 |
| Bloomz (instruction-tuned) | 7B | **35.55** | 28.57 | 29.36 | 57.71 | **40.52** | **53.11** | **40.80** |
| OpenHathi (Llama-2 CPT) | 7B | 27.69 | **30.54** | 25.51 | 57.04 | 39.03 | 32.66 | 35.41 |
| Airavata (instruction-tuned OpenHathi) | 7B | **30.43** | **29.53** | 25.60 | 55.59 | 39.04 | 41.44 | 36.93 |
| xGLM | 7.5B | 26.27 | **30.52** | 21.40 | **58.70** | **45.74** | 22.33 | 34.16 |
| Llama-3 | 8B | **40.05** | **35.48** | 27.48 | **63.07** | **45.30** | **49.33** | **43.45** |
| Aya23 | 8B | **33.68** | **36.24** | 29.88 | **64.39** | **47.18** | **52.55** | **43.98** |

* Trained for the same number of steps.

Table 10: Zero-shot evaluation of LLMs for cross-lingual language transfer in Hindi. All benchmarks report Accuracy. Models that performed better than our models have been underlined and bold, the best performance of our model has been **bold**.

| Models | Size | MMLU-Marathi | ARC-Marathi | Belebele-Marathi | Average |
|---|---|---|---|---|---|
| Paramanu-Sanskrit (ours) | 139M | 24.96 | 26.49 | 24.33 | 25.26 |
| mParamanu (ours) | 92M[*] | 23.37 | 27.53 | 25.22 | 25.37 |
| mParamanu (ours) | 162M | 25.68 | 22.16 | **28.00** | 25.28 |
| mParamanu (ours) | 237M[*] | 22.73 | 24.16 | 23.67 | 23.52 |
| Paramanu-Marathi (ours) | 208M | 25.39 | 26.49 | 27.33 | 26.40 |
| Paramanu-Marathi-instruct (ours) | 208M | **25.97** | **26.94** | 27.87 | **26.93** |
| Paramanu-Hindi (ours) | 162M | 22.83 | 24.76 | 22.78 | 23.45 |
| Paramanu-Hindi (ours) | 367M | 23.78 | 24.16 | 24.66 | 24.20 |
| Paramanu-Hindi-instruct (ours) | 367M | 24.54 | 25.63 | 26.44 | 25.54 |
| Bloom | 560M | 22.78 | 24.50 | 27.00 | 24.76 |
| Bloomz (instruction-tuned) | 560M | 26.20 | 24.24 | 25.44 | 25.29 |
| xGLM | 564M | 22.53 | 21.29 | 23.11 | 22.31 |
| Llama-3.2 | 1B | 26.37 | 23.72 | **28.11** | 26.06 |
| Bloom | 1.1B | 23.93 | 25.10 | **28.33** | 25.78 |
| Bloomz (instruction-tuned) | 1.1B | 24.82 | 21.81 | 24.0 | 23.54 |
| mGPT | 1.3B | 23.26 | 22.33 | 24.33 | 23.30 |
| xGLM | 1.7B | 22.54 | 20.43 | 22.11 | 21.69 |
| Sarvam | 2B | 23.96 | 27.53 | 26.77 | 26.08 |
| Indic-Gemma-Navrasa (instruction-tuned) | 2B | 28.29 | 22.77 | 27.00 | 26.02 |
| xGLM | 2.9B | 23.10 | 19.56 | 22.55 | 21.73 |
| Llama-3.2 | 3B | **31.83** | 24.93 | **37.33** | **31.36** |
| Nemotron-Hindi | 4B | **32.32** | 23.72 | **32.22** | 29.42 |
| xGLM | 4.5B | 25.59 | 23.29 | 24.22 | 24.36 |
| Bloom | 7B | 27.30 | 25.54 | 24.00 | 25.61 |
| Bloomz (instruction-tuned) | 7B | **32.62** | **27.44** | **53.00** | **37.68** |
| OpenHathi (Llama-2 CPT) ) | 7B | 26.09 | 24.24 | 25.88 | 25.40 |
| Airavata (instruction-tuned of OpenHathi) | 7B | 26.15 | 23.90 | **29.89** | 26.64 |
| xGLM | 7.5B | 23.73 | 21.91 | 21.77 | 22.47 |
| Llama-3 | 8B | **35.47** | 25.19 | **38.66** | **33.10** |
| Aya23 | 8B | 27.80 | 22.94 | **35.33** | 28.69 |

[*] Trained for the same number of steps.

Table 11: Zero-shot evaluation of LLMs for cross-lingual language transfer in Marathi. All benchmarks report Accuracy. Models that performed better than our models have been underlined and bold, the best performance of our model has been **bold**.

| Models | Size | Belebele-Tamil | XCOPA-Tamil | MMLU-Tamil | ARC-Tamil | Average |
|---|---|---|---|---|---|---|
| Paramanu-Tamil (ours) | 208M | 26.88 | **57.60** | 24.37 | 24.51 | 33.34 |
| Paramanu-Tamil-instruct (ours) | 208M | **30.22** | 56.00 | **26.95** | **26.04** | **34.80** |
| Bloom | 560M | 27.22 | 55.80 | 23.95 | 25.57 | 33.13 |
| Bloomz (instruction-tuned) | 560M | 23.55 | **58.60** | 25.78 | 25.30 | 33.30 |
| xGLM | 564M | 22.77 | 56.20 | 22.74 | 20.57 | 30.57 |
| Llama-3.2 | 1B | 28.44 | 55.60 | 25.95 | 21.27 | 32.81 |
| Bloom | 1.1B | 25.77 | 57.00 | 24.67 | 24.34 | 32.94 |
| Bloomz | 1.1B | 22.66 | 57.40 | 26.10 | 22.59 | 32.19 |
| mGPT | 1.3B | 20.88 | 53.20 | 23.50 | 21.36 | 29.73 |
| xGLM | 1.7B | 21.88 | 55.00 | 23.66 | 21.71 | 30.56 |
| Sarvam | 2B | 27.44 | **63.00** | 24.06 | **26.53** | **35.25** |
| Indic-Gemma-Navrasa (instruction-tuned) | 2B | 25.44 | **59.00** | **27.84** | 22.67 | 33.73 |
| xGLM | 2.9B | 23.44 | 54.20 | 24.33 | 20.84 | 30.70 |
| Llama-3.2 | 3B | **34.00** | **59.00** | **29.47** | 23.90 | **36.59** |
| xGLM | 4.5B | 22.66 | 55.20 | 24.03 | 21.19 | 30.77 |
| Bloom | 7B | 25.55 | **59.20** | 26.39 | 24.69 | 33.95 |
| Bloomz (instruction-tuned) | 7B | **50.66** | 57.40 | **29.48** | **28.10** | **41.41** |
| xGLM | 7.5B | 22.44 | 54.40 | 24.39 | 21.71 | 30.73 |
| Llama-3 | 8B | **38.55** | **59.80** | **31.66** | **26.35** | **39.09** |
| Aya23 | 8B | **33.55** | 55.60 | 26.14 | 21.19 | 34.12 |

Table 12: Zero-shot evaluation of LLMs in Tamil script models. All benchmarks report Accuracy. Models that performed better than our models have been underlined and bold, the best performance of our model has been **bold**.

| Models | Size | Belebele-Telugu | XStoryCloze-Telugu | MMLU-Telugu | ARC-Telugu | Average |
|---|---|---|---|---|---|---|
| Paramanu-Telugu (ours) | 208M | 26.00 | 51.42 | 25.12 | **26.32** | 32.22 |
| Paramanu-Telugu-instruct (ours) | 208M | **27.50** | **58.00** | **26.75** | 25.75 | **34.50** |
| Bloom | 560M | 23.55 | 55.65 | 24.10 | 23.85 | 31.78 |
| Bloomz (instruction-tuned) | 560M | 22.44 | 54.86 | **26.82** | 24.91 | 32.25 |
| xGLM | 564M | 25.11 | 55.85 | 22.91 | 17.54 | 30.35 |
| Llama-3.2 | 1B | **28.44** | 54.86 | 25.40 | 19.82 | 32.13 |
| Bloom | 1.1B | 26.88 | 56.38 | 24.53 | 24.38 | 33.04 |
| Bloomz (instruction-tuned) | 1.1B | 23.11 | 55.32 | 25.49 | 18.77 | 30.67 |
| mGPT | 1.3B | 22.88 | 57.25 | 25.85 | 17.89 | 30.96 |
| xGLM | 1.7B | 23.66 | **58.23** | 24.34 | 18.24 | 31.12 |
| Sarvam | 2B | **27.66** | **60.09** | 24.67 | 25.78 | **34.55** |
| Indic-Gemma-Navrasa (instruction-tuned) | 2B | 26.55 | **58.57** | **28.58** | 20.52 | 33.55 |
| xGLM | 2.9B | 22.66 | **60.09** | 23.45 | 18.85 | 33.51 |
| Llama-3.2 | 3B | **31.55** | **58.17** | **29.47** | 20.26 | **34.86** |
| xGLM | 4.5B | 23.66 | 57.04 | 24.87 | 19.38 | 31.24 |
| Bloom | 7B | 24.66 | 57.37 | 26.62 | 24.47 | 33.28 |
| Bloomz (instruction-tuned) | 7B | **43.11** | **58.23** | **29.55** | **27.98** | **39.71** |
| xGLM | 7.5B | 24.66 | **60.22** | 23.90 | 18.15 | 31.73 |
| Llama-3 | 8B | **36.88** | **63.53** | **32.74** | 21.22 | **38.59** |
| Aya23 | 8B | **28.55** | 54.07 | 19.91 | 21.57 | 31.02 |

Table 13: Zero-shot evaluation of LLMs in Telugu script models. All benchmarks report Accuracy. Models that performed better than our models have been underlined and bold, the best performance of our model has been **bold**.

**Bangla.** From Table 9, we see that Paramanu-Bangla 108M outperformed 10 of 13 LLMs in the range of 500M and 3B, tying with Llama-3.2 1B, rivaling Sarvam 2B despite being smaller in size by 20×. While it trails 5 of 7 LLMs above 3B, it exceeds xGLM 7.5B and Bloom 1.1B in average score across MMLU, ARC, and Belebele benchmarks, despite being smaller by 70 times than xGLM 7.5B and pretrained on only 26.21 billon tokens. Trained solely on Bangla literature, its limited domain coverage explains weaker MMLU results. Instruction-tuning on 27k Bangla instructions yields PARAMANU-BANGLA-INSTRUCT 108M, which surpasses all 13 LLMs in the 500M-3B range and 3 of 7 above 3B, including xGLM 4.5B, Bloom 7B, and xGLM 7.5B. On ARC, it outperforms all models. The multilingual variant (MPARAMANU) underperforms its monolingual counterpart, reflecting the multilinguality trade-off. Notably, Bloom 1.1-Instruct also shows a 1.3% drop in Bangla performance.

**Devanagari.** Table 11 and Table 10 show that PARAMANU models achieve strong results on Devanagari benchmarks despite their small size and limited pretraining. PARAMANU-MARATHI 208M outperforms all LLMs in the 500M-3B range and 4 of 10 models above 3B on Marathi. MPARAMANU 162M and monolingual PARAMANU-SANSKRIT, despite not being trained on Hindi and Marathi data, surpass the random baseline and 8 of 21 LLMs via cross-lingual transfer. PARAMANU-HINDI-INSTRUCT 367M, tuned on 27k Hindi instructions and 52k synthetic Alpaca dataset, exceeds 18 of 21 LLMs on Marathi, and outperforms 8 models (500M–3B) and 6 of 10 above 3B on Hindi. Though Nemotron-Hindi 4B and Aya23 8B lead in Hindi, our models are significantly more efficient. Unlike costly continual pretraining and fine-tuning of LLMs (e.g., OpenHathi 7B, Airavata 7B, Indic-Gemma 2B), our pretrain from scratch approach of tiny monolingual models for low-resource languages with language-specific tokenization yields competitive results using just 240 GPU-hours.

**Tamil.** From Table 12, our model, Paramanu-Tamil (208M), outperformed 9 of 11 in range 500M-3B including Bloom, Llama-3.2, mGPT, xGLM) across four benchmarks in Tamil, coming close to Sarvam (2B) despite being 10 times smaller and trained on 76 times less Indian tokens compared to Sarvam 2B. However, its performance on MMLU is lower than the random baseline like many LLMs including Sarvam 2B in comparison as shown in the table as the model is mostly trained on Tamil news corpora. Paramanu-Tamil-instruct which is instruction-tuned on our translated 23,000 instructions dataset performed better than all models except Sarvam 2B in the range between 500M-3B and 4 LLMs out of 7 in the range of 3B and 8B.

**Telugu.** From Table 13, Paramanu-Telugu 208M outperformed 6 models out of 11 and 3 mod-

els (XGLM 4.5B, 7.5B, Aya23 8B) between 3B and 8B despite being trained Telugu pretrained on 39.32 billion tokens. After instruction-tuning on 23,000 machine translated instructions, Paramanu-Telugu-instruct (208M) outperformed all models between 500M and 3B except Sarvam 2B in the range of 500M and 3B and underperformed than 3 models of 7 in the range of 3B and 8B. On ARC benchmark, our models peformed better than all models except Bloomz 7B. The improvements in metric scores for Tamil and Telugu instruction-tuned models were modest, likely due to lower-quality machine translations from Bangla compared to Hindi.

### C.2 n-shot degradation

In this subsection, we present Table 14, which shows n-shot performance (0-shot, 5-shot, and 25-shot) on language benchmarks.

### C.3 Human Evaluation

Automated evaluation metrics may overlook significant qualitative enhancements, especially when model outputs align well with particular linguistic or cultural contexts (Barnett et al., 2024). Figure 4 shows average human ratings for models text generation across grammar, coherence, creativity, and factuality. We perform human evaluation for Bangla and Hindi for all pretrained models. For human evaluation, we asked 10 annotators to evaluate top-3 responses for 10 prompts on a scale of 0 (worst) to 5 (best). We reached inter-annotator kappa score of 0.85 for Bengali, 0.79 for Hindi, and 0.72 for Sanskrit. Figure **??** shows the bar chart for inter-annotator agreement's Fleiss Kappa score.

### C.4 Perplexity, MFU, CPU Inference Speed

Table 15 lists the test perplexity, MFU metric during pretraining and CPU inference speed of our various pretrained models. In terms of quantitative evaluation of language modeling, the lower the perplexity, the better is the language model.

| Model | CPU Inference Speed | Perplexity | MFU |
|---|---|---|---|
| Bangla 87.25M | 42.327 | 5.069 | 35.36 |
| Bangla 108.5M | 37.351 | 4.102 | 22.57 |
| Hindi 162M | 34.492 | 16.992 | 39.45 |
| Hindi 367.5M | 12.906 | 11.052 | 39.87 |
| mParamanu 92.63M | 39.459 | 8.443 | 39.57 |
| mParamanu 162M | 34.711 | 6.924 | 39.93 |
| Marathi 207.73M | 24.875 | 8.943 | 19.50 |
| Sanskrit 139.33M | 38.676 | 1.748 | 37.81 |
| Tamil 207.84M | 24.535 | 7.618 | 18.77 |
| Telugu 208.25M | 24.125 | 5.400 | 40.07 |

Table 15: CPU Inference speed (tokens/sec, FP32), perplexity of our models and MFU metric during training. MFU is Model FLOPs Utilization metric.

## D Model & Training Configuration

In this section, we provide details on the various model architecture and pretraining hyperparameters used in our experiments. Tables 2 and 16 list these configurations. Figure 5 shows GPU utilization during the pretraining of the Hindi model on a single GPU. Figure 6 presents plots of training loss versus training steps and training loss versus training tokens for several of our pretrained models.

| Model | Params | Batch | Grad Acc. | Seq Len | LR |
|---|---|---|---|---|---|
| Bangla | 87.25M | 32 | 8 | 1024 | 0.002 |
| Bangla | 108.5M | 32 | 8 | 1024 | 0.003 |
| Hindi | 162M | 32 | 8 | 1024 | 0.002 |
| Hindi | 367.5M | 32 | 16 | 1024 | 0.003 |
| Marathi | 207.73M | 32 | 8 | 1024 | 0.003 |
| mParamanu | 92.63M | 32 | 16 | 1024 | 0.002 |
| mParamanu | 162M | 32 | 8 | 1024 | 0.002 |
| Sanskrit | 139.3M | 64 | 8 | 1024 | 0.003 |
| Tamil | 207.84M | 32 | 8 | 1024 | 0.002 |
| Telugu | 208.25M | 16 | 16 | 1024 | 0.003 |

Table 16: Training hyperparameters for various Paramanu models. All models are pretrained for 100K training steps except Hindi 367M (150K).

## E Model Outputs

In this section, Figure 7, 8, 9, 10 show various ChatGPT outputs when prompted in Sanskrit and Bangla. Table 17 shows the output of multilingual mParamanu-162M pretrained model when prompted in various Indo-European family of Indian languages.

| N-shot | XNLI-Hindi | XStoryCloze-Hindi | XStoryCloze-Telugu | XCOPA-Tamil |
|---|---|---|---|---|
| 0 | 33.49 | **52.42** | **56.06** | **54.00** |
| 5 | **34.04** | 51.49 | 54.67 | 52.40 |
| 25 | 33.23 | 52.02 | 55.92 | 49.80 |

Table 14: N-shot evaluation of pretrained Paramanu models across various benchmarks.

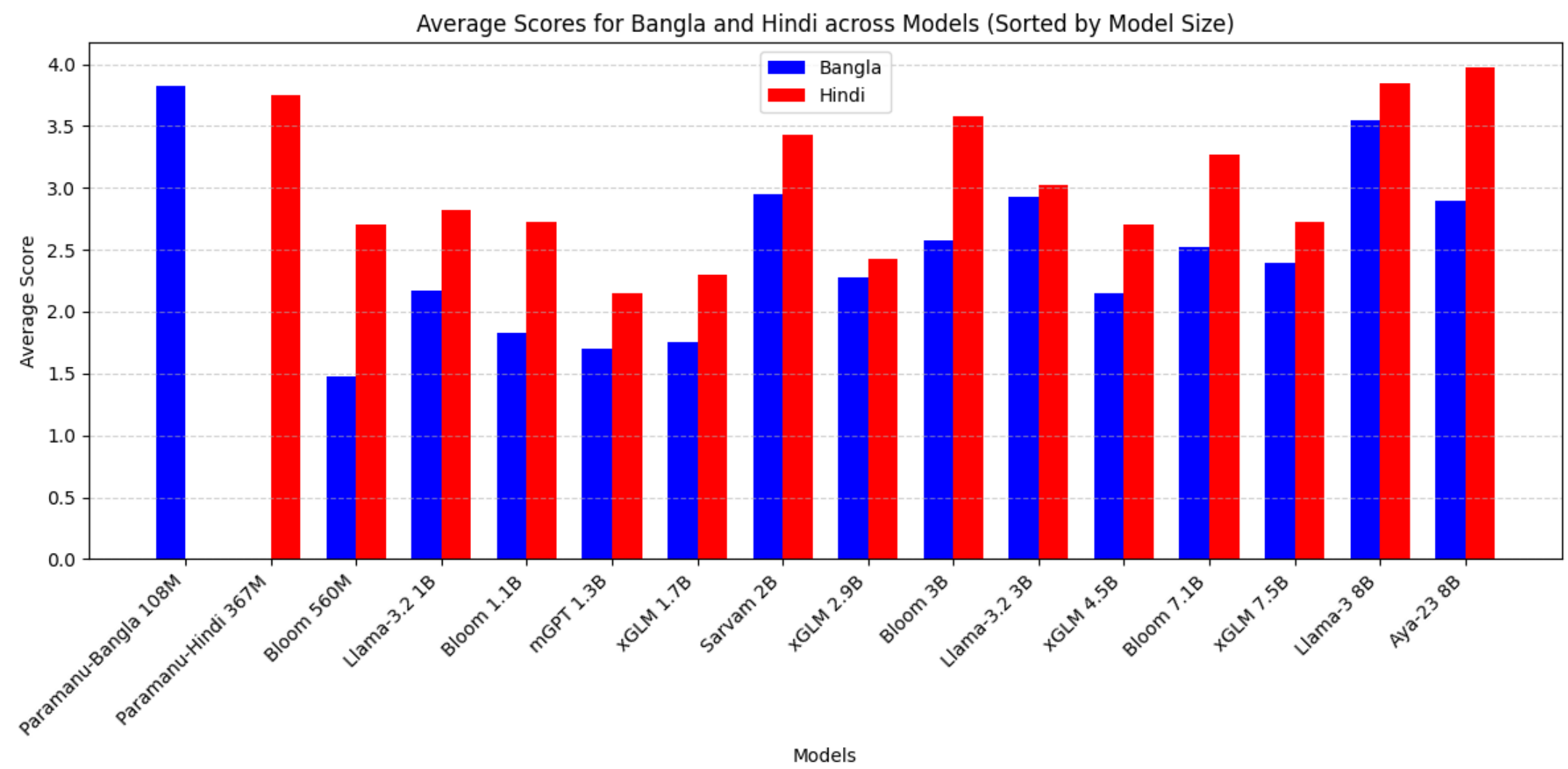

Figure 4: Human evaluation of LLMs. Average score across Grammar, Coherence, Creativity, Factuality.

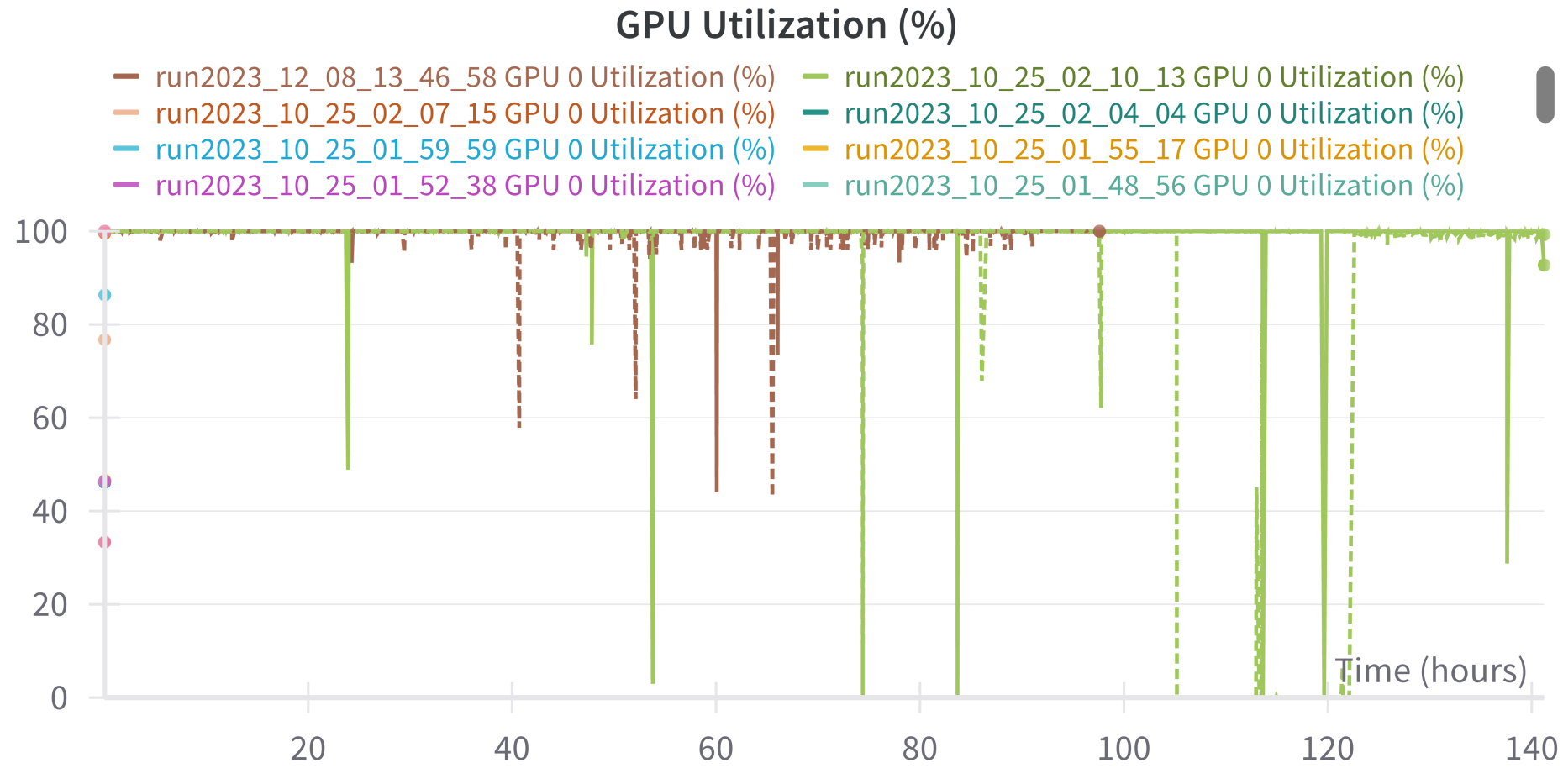

Figure 5: GPU Utilization (%) for pretraining Hindi language model from scratch. Each color indicates a run, resumed after an interruption.

| Language | Input Prompt | Text Generation |
|---|---|---|
| Assamese | ভাৰত মোৰ মাতৃভূমি | মোৰ দেশ ভাৰত বিশ্বৰ বিভিন্ন ঠাইৰ সংস্কৃতিকলৈয়ে গঢ় লৈ উঠিছে।<br>মোৰ দেশ ভাৰত এখন সাংস্কৃতিক কেন্দ্ৰ। |
| Bangla | ভারত আমার মাতৃভূমি | আমার দেশ ভারতবর্ষ আমাদের দেশ ।<br>আমার দেশ ভারতবর্ষের বিরাট অংশ এক সময় বঙ্গদেশের মত ছিল। |
| Dogri | साडा देश भारत है | साडा देश भारत है हाँ आज के समय से आने वाला उत्तर भारतीय है जहां धीरूभाई अम्बानी फॉर युवा का मुजफ्फरपुर का लोग है वहां के बावजूद बिग बॉस नाम की तरफ से भी कुछ लकी। |
| Konkani | भारत हो म्हजो देश | भारत हो म्हजो देश ह्या राष्ट्राची वाट लायल्या<br>भारत हो म्हजो देश भारत म्हज्या इतिहासांत देव म्हज्या इतिहासांत आस्पावता आदलें पैकूच आसा त्या इतिहासाचो आदलो काळ आशिल्लो हाका इतिहासाचो चवथो भाग म्हण्टा। |
| Maithili | भारत हमर देश अछि | भारत हमर देश अछि और ओ देशक सर्वोच्च पदसँ ओतप्रोत अछि<br>भारत हो म्हजो देश हो मोलांत। |
| Odia | ଭାରତ ହେଉଛି ମୋର ମାତୃଭୂମି | ଭାରତ ହେଉଛି ମୋର ମାତୃଭୂମି ଜାତିର ଏକତ୍ରୀକରଣ ହୋଇଛି<br>ଭାରତ ହେଉଛି ମୋର ମାତୃଭୂମି ଓ ମୋ ନିଜସ୍ୱତା |
| Sanskrit | अस्माकं देशः भारतवर्षः | अस्माकं देशः भारतवर्षः सर्वातिशायी।<br>अस्माकं देशः भारतवर्षः इति ख्यातः। |

Table 17: mParamanu 162 M pretrained model text generation

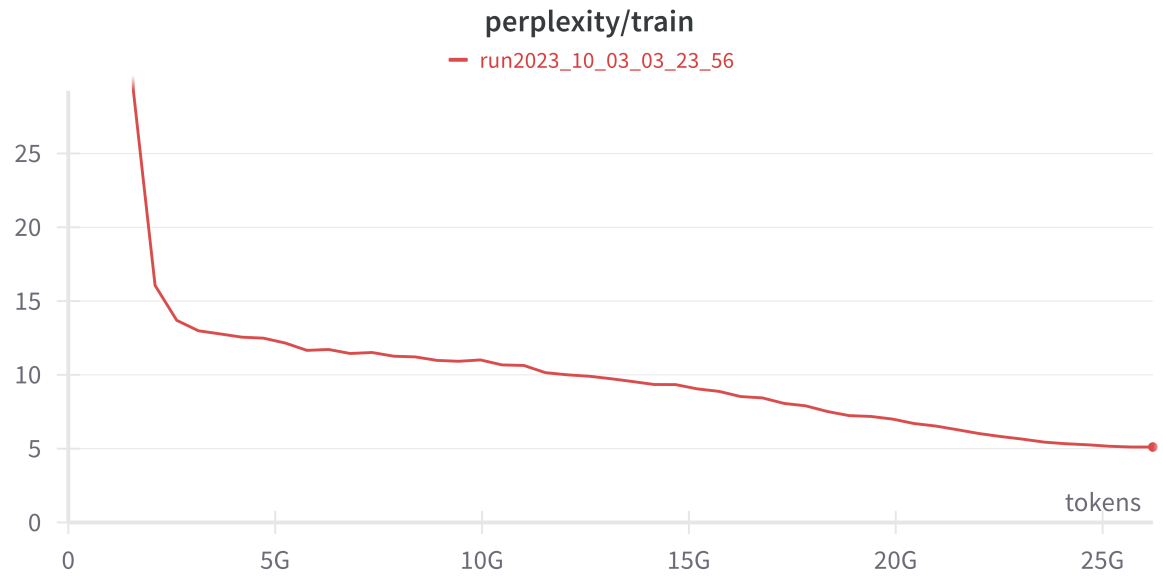

(a) Training Perplexity for Paramanu-Bangla v/s Tokens in Billion (G=billion). Each color indicates a run, resumed after an interruption.

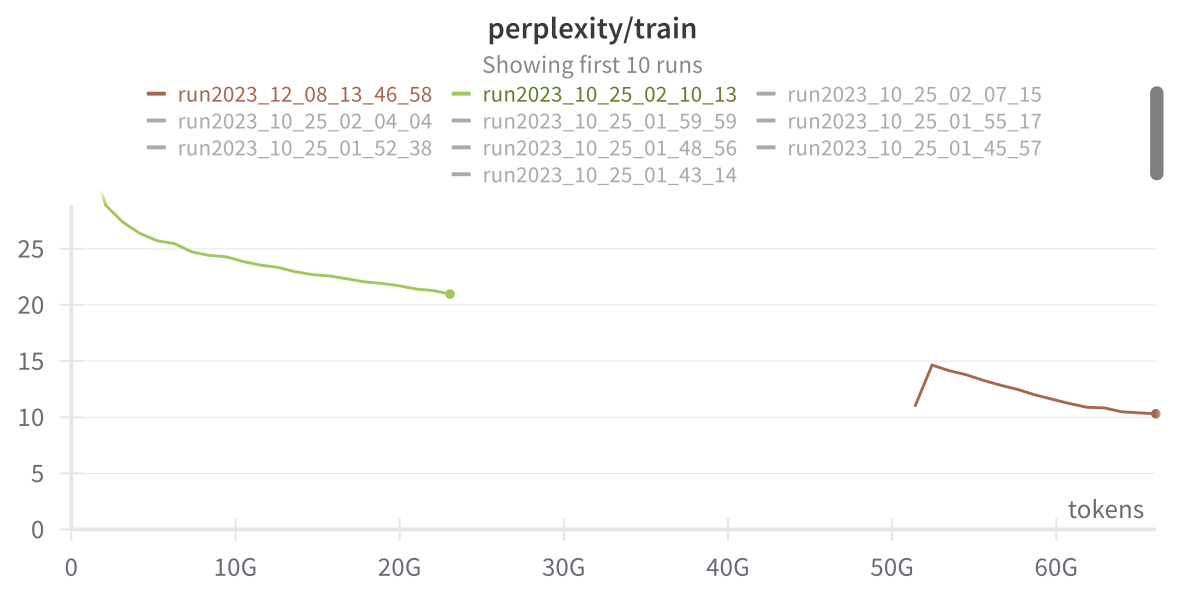

(b) Training Perplexity for Paramanu-Hindi v/s Tokens in Billion (G=billion). Each color indicates a run, resumed after an interruption.

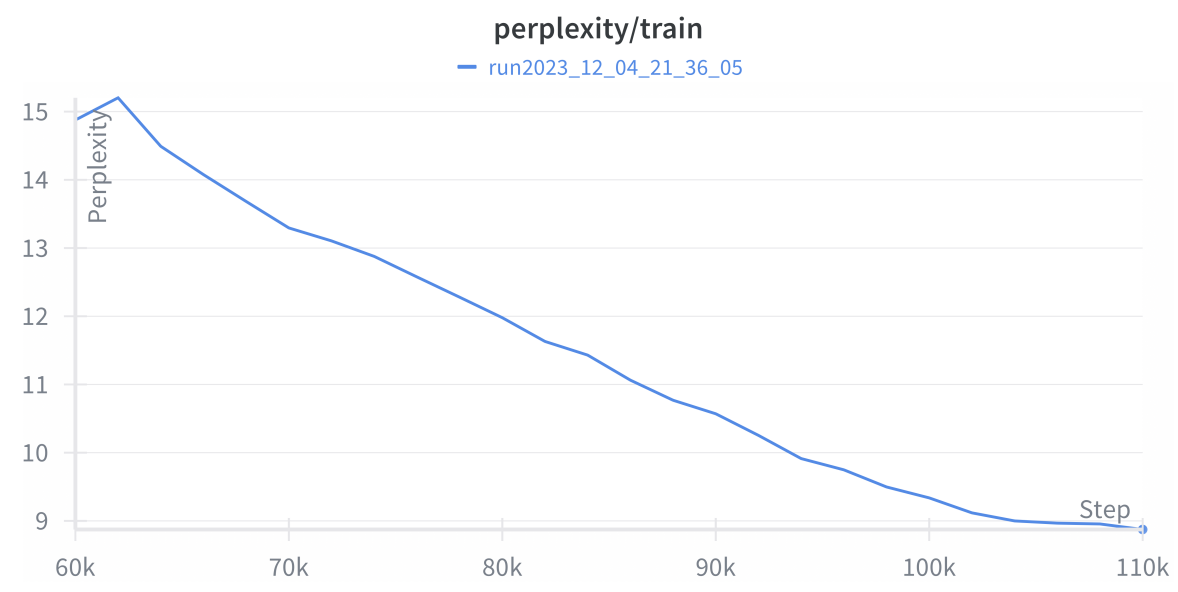

(c) Training Perplexity of Paramanu-Marathi v/s training steps. Each color indicates a run, resumed after an interruption.

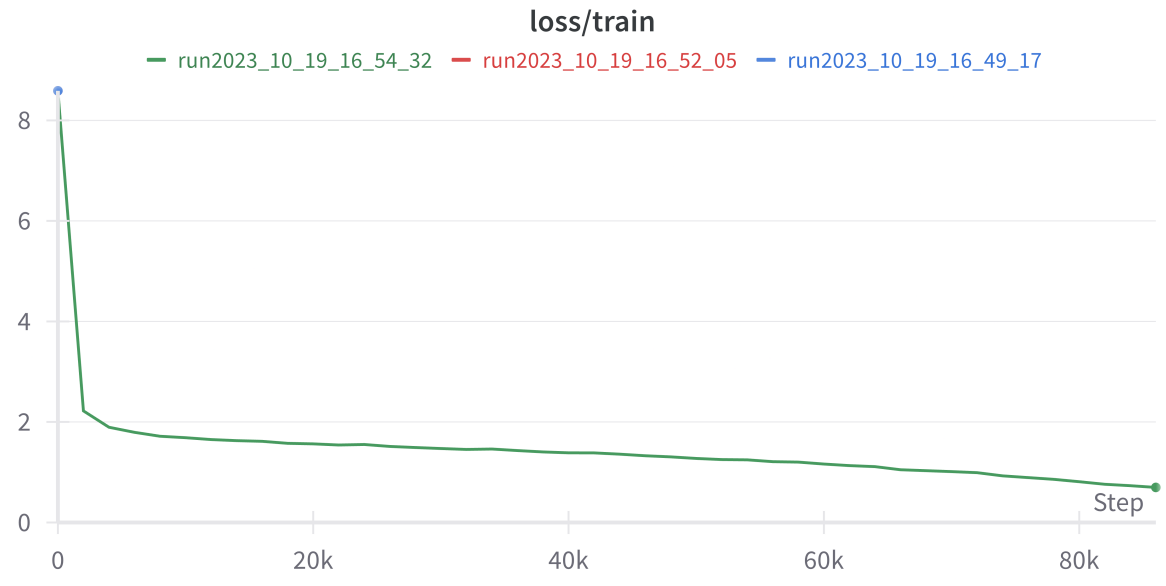

(d) Training Perplexity of Paramanu-Sanskrit v/s training steps. Each color indicates a run, resumed after an interruption.

Figure 6: Training Perplexity v/s Tokens and Training Perplexity v/s training steps for Paramanu pretrained models. Each color indicates a run, resumed after an interruption.

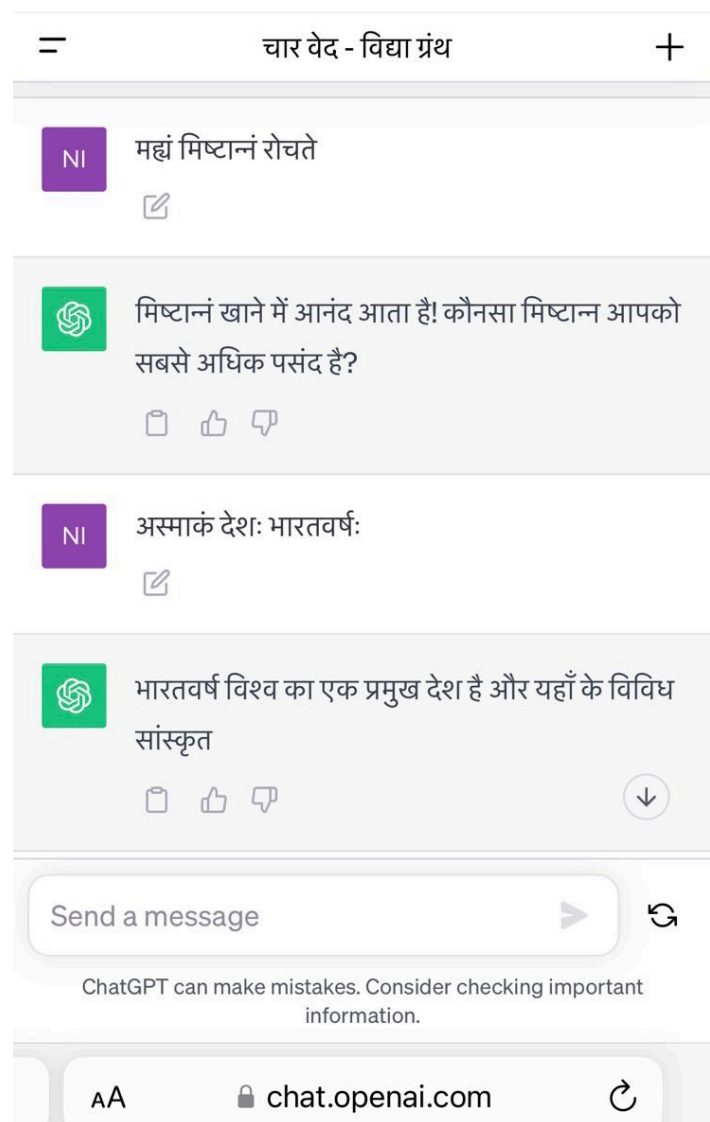

Figure 7: GPT-3.5 Turbo response to Sanskrit prompt: अस्माकं देशः भारतवर्षः (asmākaṁ dēśaḥ bhāratavarṣaḥ), मह्यं मिष्टान्नं रोचते (mahyaṁ miṣṭānnaṁ rōcatē.) Accessed in October 2023.

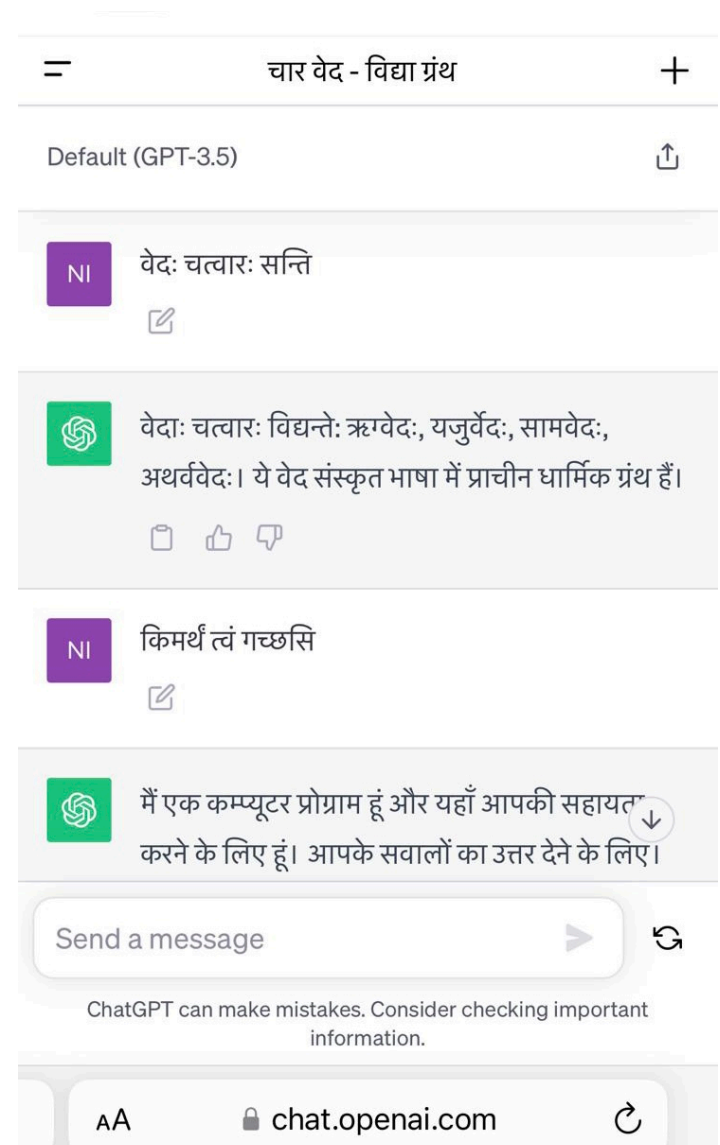

Figure 8: GPT-3.5 Turbo response to Sanskrit prompt: वेदः चत्वारः सन्ति (vēdaḥ catvāraḥ santi), किमर्थं त्वं गच्छसि (kimarthaṁ tvaṁ gacchasi.) Accessed in October 2023.

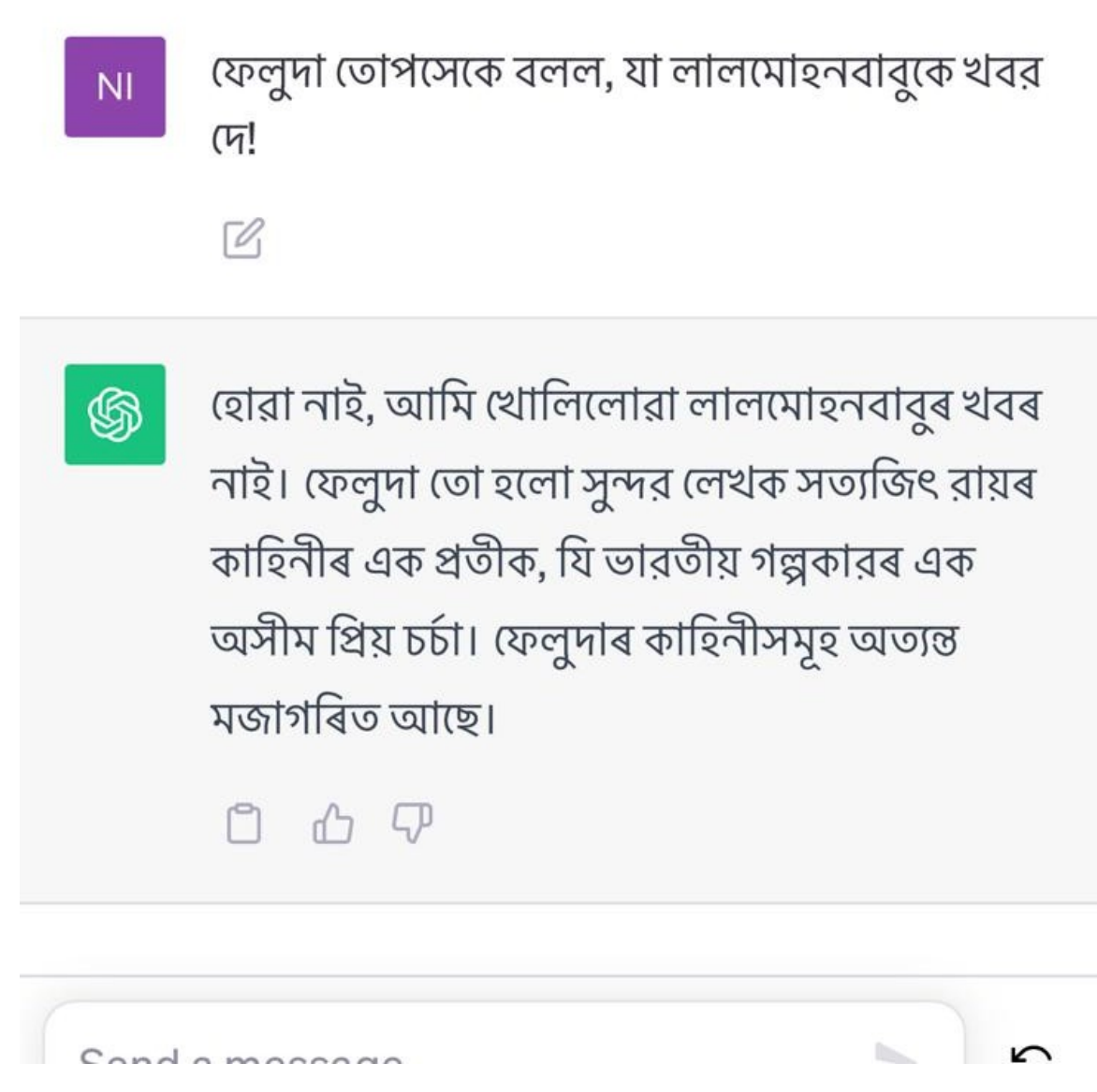

Figure 9: GPT-3.5 Turbo response to Bangla prompt ফেলুদা তোপসেকে বলল, যা লালমোহনবাবুকে খবর দে! Accessed in October 2023.

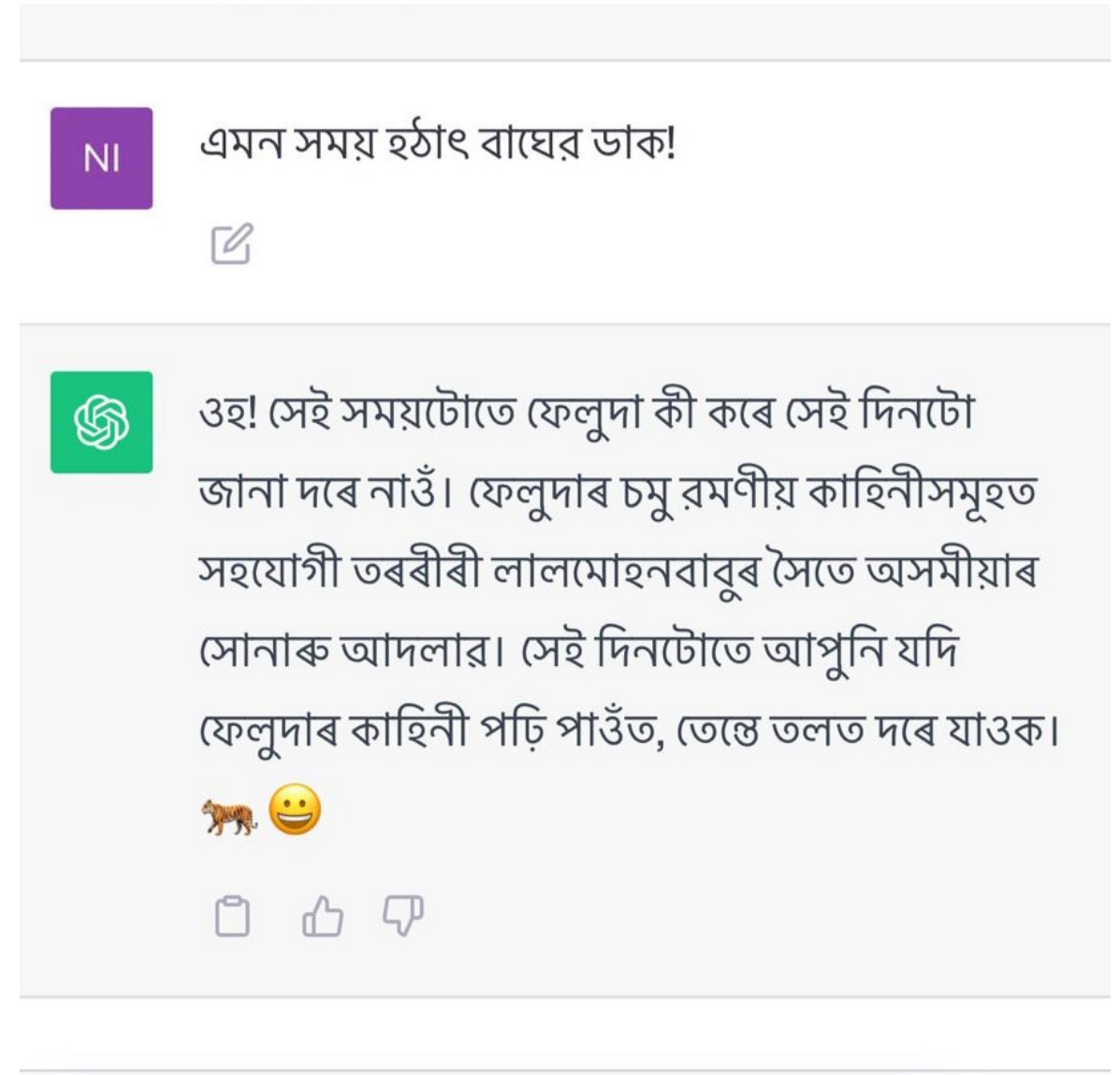

Figure 10: GPT-3.5 Turbo response to Bangla prompt এমন সময় হঠাৎ বাঘের ডাক! Acessed in October 2023.